\pdfoutput=1

\documentclass[11pt]{article}

\PassOptionsToPackage{table}{xcolor}   
\RequirePackage{xcolor}

\usepackage[final]{acl}

\usepackage{times}
\usepackage{latexsym}

\usepackage{amsmath}
\usepackage{listings}
\usepackage{amssymb}
\usepackage[T1]{fontenc}

\usepackage[utf8]{inputenc}

\usepackage{microtype}

\usepackage{inconsolata}

\usepackage{graphicx}

\definecolor{ForestGreen}{RGB}{34,139,34}

\definecolor{TealGreen}{RGB}{0,128,128}

\definecolor{InkGreen}{RGB}{0,100,0}

\definecolor{AcademicBlue}{RGB}{40,100,200}

\title{Efficiently Selecting Response Generation Strategy by Self-Aligned Perplexity for Fine-Tuning LLMs}

\author{
  Xuan Ren\thanks{Equal contribution.} \\
  University of Adelaide \\
  \texttt{xuan.ren@adelaide.edu.au} \\
  \And
  Qi Chen\textsuperscript{*} \\
  University of Adelaide \\
  \And
  Lingqiao Liu\textsuperscript{1}\thanks{Corresponding author.} \\
  University of Adelaide \\
  \texttt{lingqiao.liu@adelaide.edu.au} \\
}

\begin{document}
\maketitle
\begin{abstract}

Fine-tuning large language models (LLMs) typically relies on producing large sets of input-output pairs. Yet for a given question, there can be many valid outputs. In practice, these outputs are often derived by distilling knowledge from teacher models, and they can vary depending on the specific teacher model or prompting strategy employed.
Recent findings show that \emph{how} these training outputs are generated can significantly affect the performance of the fine-tuned model, raising an important question: how do we pick the best \emph{data generation method} from among numerous possibilities? Rather than exhaustively training and evaluating on each candidate, this paper proposes a scalable approximate method that assesses a \emph{small} subset of generated data to estimate its suitability for a specific target LLM. Our central idea is that effective outputs should be \emph{familiar} to the target LLM. While previous work measures familiarity with perplexity, we find that perplexity might be suboptimal in characterizing ``familiarity'' through empirical analyses and practical observations. To address this, we introduce \emph{self-aligned perplexity}, a novel metric capturing how closely candidate outputs adhere to the target LLM’s own style and reasoning patterns. In this way, we can identify the most effective generation strategy on a small sample, then apply it to produce the complete training set. We demonstrate that training on data generated by the chosen method yields significant improvements across diverse reasoning-focused benchmarks, particularly in cases where different candidate methods lead to highly divergent training outcomes. Our implementation is publicly available at \url{https://github.com/XuanRen4470/SPPL}.

\end{abstract}

\section{Introduction}

When instruction-tuning an LLM, training data consists of question-response pairs, where multiple valid responses can be generated for the same input. 
Previous studies ~\citep{ren2024learn} show that datasets with identical input questions but different responses can lead to varied learning outcomes, even when responses contain similar levels of detail.
This raises a key question: \textit{how can we construct responses that are most effective for the target LLM?}

Prior research has explored improving responses by adding details or rationales, such as structuring ground truth step by step~\citep{hsieh2023distilling, ranaldi-freitas-2024-aligning}, incorporating rationales, or enriching responses with additional information~\citep{zhang2024distillation, kang2023knowledgeaugmented, li2022explanations}. However, recent studies~\citep{ren2024learn, yang-etal-2024-self} suggest that more details or converting responses to step by step style do not always improve performance and that alignment with the LLM’s linguistic style is crucial.

In our experiment, we observe that no single response generation strategy works universally across tasks. Thus, we need to create a method to find out the most effective way to generate responses for each task, rather than a single method for all tasks.

The concurrent works \citep{xu2024strongermodelsstrongerteachers, kim2024evaluatinglanguagemodelssynthetic} attempt to predict the effectiveness of response generation methods by evaluating the entire training dataset. They generate full training datasets using each method and then estimate training effectiveness based on scores computed via algorithms or reward models. However, these approaches are computationally expensive and not scalable.

However, can we predict the effectiveness of each data generation methods efficiently? We observe an interesting phenomenon that each response generation method produces responses with a consistent style, meaning that a small subset of generated examples can effectively represent the entire dataset. Based on this assumption, we propose an efficient ranking pipeline that evaluates a limited number of samples (e.g., 50) to assess the performance of each response generation strategy. This approach uses an alignment estimation function to assign scores to each strategy, enabling us to identify the best-performing method without the need for a full-dataset evaluation. 

Previous research \citep{ren2024learn} used perplexity to measure a model’s familiarity with candidate question-answer pairs, proposing that lower-perplexity responses for the same input tend to yield better training performance. However, we found several cases where perplexity-based filtering was ineffective. For instance, responses structured in a step-by-step or redundant style often exhibit low perplexity but do not necessarily improve training outcomes. While some candidate responses (e.g., step-by-step or redundant ones) may achieve low perplexity under the target LLM, the model itself rarely generates such responses when producing answers freely. This suggests that low perplexity does not always indicate alignment with the model’s inherent reasoning style. These findings suggest that perplexity can be "hacked" by response style. Thus, traditional perplexity alone is insufficient for selecting the best response generation strategy.

To address this, we propose self-aligned perplexity, a refined metric for measuring a model’s familiarity with target responses. The key idea is that a model is most familiar with the data it generates itself. Leveraging this, we modify perplexity computation by incorporating model-generated responses as in-context examples. Specifically, we first have the model produce initial responses, which is then appended to the question as in-context examples. A prompt enforce the model to pay attention to these examples when computing perplexity, thereby altering the probability estimation of the candidate response. If the target response deviates significantly from the model’s own generated response—the one it is most familiar with—the model assigns it a lower probability, increasing its perplexity. Our experiments show that self-aligned perplexity outperforms traditional perplexity in selecting effective data generation strategies.

In our experiments, we observe a strong correlation between the proposed indicator and the ranking of training dataset performance. Furthermore, we construct a pool of answer generation strategies and demonstrate that applying our selection criterion leads to significant performance gains compared to the baselines—especially in scenarios where different data-generation methods produce highly divergent outcomes

\section{Related Works}
%
There has been extensive research into what types of data yield the best training outcomes for large language models (LLMs). Previous studies have identified several factors that positively influence model training, such as adding complexity \citep{xu2023wizardlm}, adding details \citep{zhang2024distillation, kang2023knowledgeaugmented, li2022explanations}, adding diversity \citep{luo2023wizardcoderempoweringcodelarge}, augmenting ground-truth answers in a step-by-step manner \citep{hsieh2023distilling, ho2022large, magister-etal-2023-teaching, fu2023specializing, ranaldi-freitas-2024-aligning}, and ensuring correctness \citep{trinh2024solving, ranaldi-freitas-2024-aligning}. However, in practice, these metrics are challenging to measure for a given dataset, making it difficult to determine the quality of training data based on these criteria. \citet{ren2024learn} found that familiarity, measured by perplexity, significantly impacts model training. 

Perplexity has been widely used for different purpose in prior research. Perplexity has been used to select prompts \citep{gonen2022demystifying}, showing that prompts with lower perplexity generally lead to better performance in question-answering tasks. It has also been used for selecting pretraining datasets \citep{de2022bertin}, detecting AI-generated content \citep{xu2024detecting, hu2020systematic}, and selecting instruction-tuning data from a database \citep{mekala2024smaller}. \citet{li-etal-2024-quantity} modify the perplexity score and propose “IFD” (Instruction Following Difficulty), which is used to select a small pool of challenging data from the original dataset for efficient training. Researchers hypothesize that higher perplexity indicates more challenging data, which can be beneficial for teaching LLMs new knowledge. In addition, perplexity or confidence-based curricula have been explored for NMT \citep{kocmi2017curriculum} and general text generation \citep{platanios2019competence}, where harder (high-perplexity) data are introduced progressively to improve sample efficiency. Unlike these studies, our focus is on identifying the best strategy to generate target responses (y) for a given input (x), rather than selecting difficult (x, y) pairs for training language models.

Recent efforts have begun to ask which teacher model produces the most useful synthetic targets. \citet{xu2024strongermodelsstrongerteachers} introduce a \textit{Compatibility-Adjusted Reward} (CAR) and judge its quality by the Spearman correlation between CAR scores and downstream accuracy on two instruction-following datasets, each evaluated with a single meta-prompt. \citet{kim2024evaluatinglanguagemodelssynthetic} study nine datasets spanning mathematics, coding, and general instructions; they correlate several corpus statistics with training gains and combine them with principal-component analysis to rank teacher models. Our study differs in four key respects. First, we estimate a strategy’s quality from only a small sample of its outputs, making synthetic data generation and evaluation far more affordable. Second, we target accuracy improvement, not just rank correlation. Third, we experiment on a much broader benchmark: 17 diverse tasks plus 6 Plan-Bench planning tasks. Fourth, we evaluate our method on datasets generated using diverse meta-prompts, explicitly accounting for prompt variability.

\section{Method}
This paper aims to efficiently select the most effective answer generation strategy for fine-tuning a target LLM. In what follows, we first present the problem setup, then detail our proposed \emph{self-aligned perplexity} metric for scoring the outputs from each candidate strategy.


\subsection{Problem Definition}\label{sec:problem}

Let \(\mathcal{S} = \{ S_1, \dots, S_n \}\) be a set of candidate answer-generation strategies, where each strategy \(S_k\) produces a response \(\hat{y}^k = S_k(x)\) for an input \(x\). Our goal is to select the strategy \(S_\iota\) that yields the most effective training data \(\mathcal{D} = \{(x, \hat{y}^k)\}\) to fine-tune a target model \(M\). Since generating the full dataset via the API for every strategy is costly, we evaluate a small subset \(\mathcal{D}_s\) of size \(K\) (\(K \ll |\mathcal{D}|\)) to estimate how well each strategy’s outputs align with \(M\).

\subsection{The Familiarity Hypothesis}
The work in \cite{ren2024learn} suggests that if the model is more ``familiar'' with a given response, then the model can learn better with the given response. In their work, perplexity, which is correlated to the likelihood of generating a response with the model, is used to measure this familiarity score. In our study, we argue that perplexity is sub-optimal to measure familiarity. We suggest that familiarity can be more precisely measured by this equation:
\begin{align}
    F(\hat{y}) = \mathbb{E}_y \left[s(y,\hat{y})\right] = \int s(y,\hat{y}) P_M(y) dy,
\end{align}
where $s(y,\hat{y})$ is a semantic similarity measure between $\hat{y}$ and a sample response $y$ drawn from the model $M$. In plain language, it quantifies how similar a candidate response is to the range of answers that the model might generate. It is straightforward to demonstrate that when $s(y,\hat{y}) = \delta(y,\hat{y})$, i.e., when $\delta(y,\hat{y}) = 1$ only if $y$ is exactly identical to $\hat{y}$, the function $F$ becomes equivalent to the likelihood $P_M(\hat{y})$, and hence equivalent to perplexity. using perplexity as a surrogate to measure familiarity fails to account for the variety of responses that may be semantically equivalent to a candidate response, thereby underestimating the familiarity. In practice, this results in assigning an excessively high perplexity to a good candidate response that the model might actually be familiar with, as evidenced by our empirical study in section~\ref{sec:EmpiricalStudyOfWhyDoesSelfAlignedPerplexityWorkBetterThanTraditionalPerplexity}.

\subsection{Self-Aligned Perplexity}
\label{sec:self_aligned_perplexity}

To evaluate the effectiveness of different response generation strategies, we first construct a small calibration set $\mathcal{D}_s$, consisting of the first \(K\) examples (e.g., \(K{=}50\)) input questions. For each strategy, we generate one candidate response \(\hat{y}_i\) per question \(x_i \in \mathcal{D}_s\). Rather than directly measuring the likelihood of a response given a question, as in traditional perplexity, we compute its likelihood under a prompt augmented with in-context exemplars to better reflect stylistic alignment.

Specifically, for each input \(x_i\), we let the target model \(M\) generate an initial prediction \(y_i = M(x_i)\). Since these initial predictions may be incorrect, we evaluate them using the same task-specific script used at test time, and retain only those that are correct. These correct model-generated responses form a \emph{style pool} \(C\).

To measure how well a candidate \(\hat{y}_i\) (generated by a specific strategy) aligns with the model’s own response style, we compute its perplexity under a prompt that includes two in-context exemplars randomly sampled from the \emph{style pool} \(C\) (see Appendix~\ref{appendix:self_aligned_ppl_prompt} for the prompt). Each exemplar \(y_{s_1}, y_{s_2} \in C\) comes from a different question than \(x_i\) (\(s_1 \ne i, s_2 \ne i\)) to avoid leaking the answer.

This measurement corresponds to what we refer to as the \textit{self-aligned perplexity (SPPL)} of \(\hat{y}_i\). Compared to standard perplexity, which only conditions on the input question, SPPL incorporates stylistic guidance from the model’s own correct outputs. Lower SPPL indicates stronger alignment with the model’s preferred generation style.

For each response generation strategy \(S_k\), we compute the SPPL for all of its generated responses in \(\mathcal{D}_s\), and average the scores to obtain a strategy-level value \(\pi_{\texttt{SPPL}}(S_k)\). We then rank all strategies by their average SPPL and select the one with the lowest score to generate the full training set.

\section{Benchmark Construction}

In this section, we show how we use different strategies (distinct prompts and teacher LLMs) in generating high-quality responses with different styles.

\subsection{Target LLMs and APIs}

We use  Mistral-7B-instruct-V2 \citep{jiang2023mistral}, Llama3-instruct \citep{dubey2024llama3herdmodels} and  Qwen-2.5-7B-Instruct\citep{qwen2025qwen25technicalreport} as the target language models $M$. In this paper, we refer to Llama3-instruct, Mistral-7B-instruct-V2, and Qwen-2.5-7B-Instruct as Mistral7B, Llama3, and Qwen2.5, respectively. We use GPT-4o, MiniGPT-4o, and Claude 3.5 APIs as teacher models for response generation. Specifically, we use gpt-4o-mini-2024-07-18 and gpt-4o-2024-08-06 \citep{openai_gpt4_api} from OpenAI, and claude-3-5-sonnet-20240620 \citep{anthropic_claude_api} from Anthropic.

\subsection{Datasets}
\label{datasets}
We use English reasoning datasets referenced in the technical reports of LLaMA3 \citep{dubey2024llama3herdmodels}, Mistral \citep{jiang2023mistral}, and Qwen-2.5 \citep{qwen2025qwen25technicalreport} (the three target models \( M \) in our experiments). We select datasets with at least 650 examples that can be evaluated via accuracy. If a dataset lacks sufficient training data, we reconstruct it to contain at least 400 training, 50 validation, and 200 testing examples.

For datasets with subcategories (e.g., MATH, MMLU, MMLU\_PRO, API\_BANK, AGIEVAL), we choose the challenging subcategory (i.e., with the lowest reported accuracy). For example, we include moral scenarios from MMLU, Professional Law from MMLU\_PRO, Level 3 problems from API\_BANK, geometry from MATH, and LogicQA from AGIEVAL; we also incorporate the Algebra subcategory from MATH as in \cite{ren2024learn}.

Following \cite{ren2024learn}, we train and evaluate the first 1,000 training and testing examples, generating up to 1,000 training examples per data generation strategy.

\paragraph{Main-experiment corpus.}
\label{paragraph:Main-experiment corpus}
In total, our datasets include: \textbf{Mathematics:} GSM8K \citep{cobbe2021training}, MATH (Algebra) and MATH (Geometry) \citep{hendrycks2021measuring}; \textbf{Commonsense reasoning:} PIQA \citep{bisk2020piqa}, WinoGrande \citep{sakaguchi2021winogrande}, Hellaswag \citep{zellers-etal-2019-hellaswag}, and ECQA \citep{aggarwal2021explanations}; \textbf{Reading comprehension:} BoolQ \citep{clark2019boolq} and SQuAD \citep{rajpurkar-etal-2016-squad}; \textbf{Aggregated benchmarks:} MMLU (Moral Scenarios) \citep{hendrycks2020measuring}, MMLU\_PRO (Professional Law) \citep{wang2024mmlu}, and AGIEval (LogicQA) \citep{zhong2023agieval}; \textbf{Coding:} MBPP \citep{austin2021program}; \textbf{Reasoning:} DROP \citep{dua2019dropreadingcomprehensionbenchmark} and ARC-Challenge \citep{clark2018think}; and \textbf{Tool-using:} API-BANK (Lv 3 problems)~\citep{li2023api}.
More details are in Table~\ref{tab:why_each_data_is_chosen} (Appendix).

\paragraph{PlanBench Extension.}\label{paragraph:PlanBench extension}
We further evaluate the most challenging subtasks of PlanBench~\citep{valmeekam2023planbench}—those on which GPT-3 attains an accuracy below 20\%. 
The subtasks comprise \emph{plan generation}, \emph{plan optimization}, \emph{plan verification}, \emph{plan reuse}, \emph{plan generalization}, and \emph{replanning}. 
Although we experimented with various prompt formats, Qwen consistently failed to solve any \emph{execution} problems.
Since our method relies on generating correct responses for use as in-context examples, we exclude the execution task from our evaluation. 
The remaining six categories remain sufficiently challenging and are not part of our main training benchmarks. We use them solely to analyze performance variance when models are trained on tasks that are very challenging. How do different response generation strategies affect performance variance under such conditions?

\subsection{Data Generation Strategies}
\label{sec:data_generation_strategy}
Given 1,000 samples, we use different strategies to generate target responses.
For a fair comparison, we use the same prompts from \citep{ren2024learn} to generate responses, including \textbf{\hyperref[prompt answer directly]{GPT-4o Answer Directly}}, \textbf{\hyperref[prompt answer directly]{Claud Answer Directly}}, \textbf{\hyperref[prompt answer directly]{MiniGPT-4o Answer Directly}}, \textbf{\hyperref[prompt step by step]{Step-by-Step}} and 
\textbf{\hyperref[prompt rewrite groundtruth]{Rewrite Ground Truth}}. Besides, we design two new prompts named \textbf{\hyperref[prompt gpt4o example]{GPT-4o Examples}} and \textbf{\hyperref[prompt human written example]{Human Examples}} on our own. Please refer to Appendix~\ref{appendix:Response Construction Details} for details on each response construction method.

We provide ground truth to the teacher models and allow up to three attempts for data generation. If the first result is incorrect, we regenerate; otherwise, we stop. The same evaluation script used during testing is applied to check correctness.

\section{Experiment}

\begin{table*}[t!]
  \centering
  \resizebox{0.85\textwidth}{!}{
  \begin{tabular}{l|l|c|c|c|c|c|c}
    \hline
    Methods & STD Range & num of recorded data & mistral & llama 3 instruct & qwen & Avg Acc \\ \hline
    \hline
    Upper bound  & All Data & 51.0 & 59.39\% & 64.44\% & 71.41\% & 65.08\%  \\
    Step-by-step  &   &   & 56.25\% & 60.94\% & 69.87\% & 62.35\% \\
    GPT-4 ICL examples  &   &   & 57.29\% & 62.03\% & 69.90\% & 63.07\% \\
    Human examples  &   &   & 56.95\% & 61.91\% & 70.22\% & 63.03\%  \\
    Mini-GPT-4  &   &   & 56.72\% & 61.36\% & 70.13\% & 62.74\% \\
    GPT-4  &   &   & 56.92\% & 62.83\% & 69.89\% & 63.21\% \\
    Claude  &   &   & 57.45\% & 62.93\% & 70.30\% & 63.56\% \\
    Ours  &   &   & 58.34\% & 63.66\% & 70.50\% & 64.16\% \\
    \hline
    Ours - Claude  & All Data & 51.0  & \textcolor{ForestGreen}{+0.88}\% & \textcolor{ForestGreen}{+0.72}\% & \textcolor{ForestGreen}{+0.19}\% & \textcolor{ForestGreen}{+0.60}\% \\
    Ours - Avg of Others  &   &   & \textcolor{ForestGreen}{+1.41}\% & \textcolor{ForestGreen}{+1.66}\% & \textcolor{ForestGreen}{+0.45}\% & \textcolor{ForestGreen}{+1.17}\% \\
    \hline
    Ours - Claude  &$STD > 2.00\%$ & 21.0  & \textcolor{ForestGreen}{+1.67}\% & \textcolor{ForestGreen}{+1.14}\% & \textcolor{red}{-0.21}\% & \textcolor{ForestGreen}{+1.18}\% \\
    Ours - Avg of Others  &   &   & \textcolor{ForestGreen}{+2.17}\% & \textcolor{ForestGreen}{+2.84}\% & \textcolor{ForestGreen}{+1.88}\% & \textcolor{ForestGreen}{+2.41}\% \\
    \hline
    \hline
    Ours - Claude  &  $STD > 4.00\%$ & 10.0  & \textcolor{ForestGreen}{+3.61}\% & \textcolor{ForestGreen}{+1.56}\% & \textcolor{red}{-1.63}\% & \textcolor{ForestGreen}{+2.26}\% \\
    Ours - Avg of Others  &   &   & \textcolor{ForestGreen}{+3.60}\% & \textcolor{ForestGreen}{+6.00}\% & \textcolor{ForestGreen}{+3.38}\% & \textcolor{ForestGreen}{+4.54}\% \\
    \hline
  \end{tabular}}
\caption{Comparison of our method with other response generation strategies, averaged over three subsets. Experiments are conducted on datasets from the \hyperref[paragraph:Main-experiment corpus]{Main-experiment corpus}, introduced in Section 4.2. In this benchmark, Claude emerges as the strongest competitor among the baseline methods.}
\label{tab:icppl vs other response generation methods: average of 3 runs}
\end{table*}

\begin{table*}[t!]

  \centering
  \resizebox{0.85\textwidth}{!}{
  \begin{tabular}{l|l|c|c|c|c|c|c|c}
    \hline
    Methods & STD Range & num of recorded data & mistral & llama 3 instruct & qwen & Avg Acc & Weighted Spearman Pho \\ \hline
    \hline
    Upper bound  & All Data & 51.0 & 59.39\% & 64.44\% & 71.41\% & 65.08\% &   \\
    skywork  &   &   & 56.32\% & 61.64\% & 70.26\% & 62.74\% & 0.271 \\
    CAR  &   &   & 56.36\% & 61.80\% & 70.31\% & 62.82\% & 0.279 \\
    IFD  &   &   & 57.23\% & 61.65\% & 69.85\% & 62.91\% & 0.191 \\
    perplexity  &   &   & 57.48\% & 63.65\% & 70.37\% & 63.83\% & 0.301 \\
    Ours  &   &   & 58.34\% & 63.66\% & 70.50\% & 64.16\% & 0.324 \\
    \hline

    Ours - Perplexity  &  All Data &  51.0 & \textcolor{ForestGreen}{+0.86}\% & \textcolor{ForestGreen}{+0.00}\% & \textcolor{ForestGreen}{+0.12}\% & \textcolor{ForestGreen}{+0.33}\% & \textcolor{ForestGreen}{+0.023} \\ \hline
    Ours - Perplexity  &  STD > 2.00\% &  21.0 & \textcolor{ForestGreen}{+1.62}\% & \textcolor{ForestGreen}{+0.08}\% & \textcolor{ForestGreen}{+0.50}\% & \textcolor{ForestGreen}{+0.80}\% & \textcolor{ForestGreen}{+0.032} \\ \hline
    Ours - Perplexity  &  STD > 4.00\% &  10.0 & \textcolor{ForestGreen}{+3.05}\% & \textcolor{ForestGreen}{+0.18}\% & \textcolor{ForestGreen}{+1.63}\% & \textcolor{ForestGreen}{+1.76}\% & \textcolor{ForestGreen}{+0.075} \\

    \hline
  \end{tabular}}
\caption{We compare our method against IFD \citep{li-etal-2024-quantity}, Skywork \citep{liu2024skyworkrewardbagtricksreward}, CAR \citep{xu2024strongermodelsstrongerteachers}, and Perplexity\citep{ren2024learn}. The experiments are conducted on datasets from the \hyperref[paragraph: Main-experiment corpus]{Main-experiment corpus}, introduced in Section 4.2. In this benchmark, Perplexity emerges as the strongest competitor among the baselines.}
\label{tab:icppl vs other metrics: average of 3 runs}
\end{table*}

In this section, we treat each generation strategy from \hyperref[sec:data_generation_strategy]{Section~4.3}, and response-selection metrics from the related work section, as baselines. We then compare the average training outcomes of our method against these baselines across all tasks.

There are two benchmark sets, detailed in Section~\ref{datasets}. The first is the \hyperref[paragraph:Main-experiment corpus]{Main-experiment corpus}, which covers a diverse range of tasks and serves as the primary benchmark for evaluating both the general ranking ability of our metric and the average performance gains achievable by our method. 

Since our goal is efficient data selection, we evaluate each metric using only a small subset of the training data. For each method, we repeat the process three times, each time selecting a different subset of size K=50 from the training dataset, and report the average performance across these runs. For example, one run may use the first 50 samples, another the second 50, and so on. The final result is computed as the average of these three evaluations.

\subsection{Hyperparameters}
We utilize the identical hyperparameter settings as referenced in \citep{ren2024learn}. Specifically, for model fine-tuning, a learning rate of 2e-5, a batch size of 32, and a warm-up phase encompassing 10\% of the total training iterations are applied. A cosine annealing schedule is implemented for the learning rate, and only the Q and V matrices of the LoRA parameters are fine-tuned with a rank of 8. All models undergo training and evaluation using half-precision arithmetic.

\subsection{Evaluation Metrics}
\label{sec:evaluation_metrics}

\paragraph{Accuracy.}  
For every \{model,\,dataset\} pair, we let each ranking metric select the top-ranked response-generation strategy, fine-tune the model on data produced by that strategy, and record the resulting test accuracy.  
We then report the \emph{macro average} of these accuracies across all evaluated tasks.  
This score answers the practical question: \emph{If I trust a metric to choose my training data, how well will my model perform on average?}


\paragraph{Weighted Spearman correlation.}  
To measure how closely a metric’s ranking matches the gold ranking, we compute a weighted Spearman coefficient in which each task is weighted by the standard deviation of accuracies obtained from all candidate strategies; tasks whose choice of strategy matters more thus contribute more.  
The exact formula and implementation details are provided in Appendix~\ref{sec:appendix_weighted_spearman}.

\subsection{Comparison with Baseline Response Generation Strategies}
\label{experiment_all_data}

Table~\ref{tab:icppl vs other response generation methods: average of 3 runs} summarizes the average test accuracy obtained when the target model is fine-tuned on data produced by each response-generation strategy.
For datasets that provide chain-of-thought (CoT) groundtruth, we additionally evaluate the \textbf{\hyperref[prompt rewrite groundtruth]{Rewrite Ground Truth}} strategy. As this strategy is only applicable to CoT datasets and some datasets do not have CoT groundtruth, its results are excluded from the table to avoid skewing the overall averages; nevertheless, they are included in every metric that ranks candidate strategies on a per-task basis.

\paragraph{Effect of task-specific variance.}
Table~\ref{tab:ntrain_1000_lr_2e-05_seedaverage_of_seed_0,1,2} shows that the performance gap among generation strategies is highly task-dependent: some tasks show differences of several percentage points, while others are nearly insensitive to the chosen strategy. 
To quantify how much our method helps when the choice of generation strategy matters most, we group every \{model, dataset\} pair by the standard deviation (SD) of accuracies across baselines.
\textbf{All tasks} include all pairs without filtering.
\textbf{High-variance tasks} retain only those with $SD > 2\%$.
\textbf{Very-high-variance tasks} retain only those with $SD > 4\%$.
In the whole \hyperref[paragraph: Main-experiment corpus]{Main-experiment corpus}, our approach delivers the highest mean accuracy, exceeding the strongest single baseline (Claude) by 0.60\% and the mean of all baselines by 1.17\%.

When we restrict evaluation to high-variance tasks, the average gain of our method over Claude increases to 1.18\%; under the very-high-variance filter, this gain further rises to 2.26\%. Compared with the mean of all baselines, the improvements reach 2.41\% and 4.54\% on the high-variance and very-high-variance subsets, respectively. These results confirm that \emph{self-aligned perplexity} is particularly valuable in scenarios where candidate generation strategies produce widely divergent training outcomes.

For a more detailed analysis of performance variance across different tasks and models, please refer to Section \ref{Why Do Some Datasets Show Greater Variance in Training Outcomes?}.

\begin{table*}[t!]
  \centering
  \resizebox{1.0\textwidth}{!}{
  \begin{tabular}{l|l|c|c|c|c|c|c|c}
    \hline
    Methods & STD Range & num of recorded data & mistral & llama 3 instruct & qwen & Avg Acc & Weighted Spearman Pho \\ \hline
    \hline
    Upper bound  & All Data & 18.0 & 52.88\% & 54.87\% & 41.87\% & 49.88\% & \\
    Step-by-step  &   &   & 37.56\% & 44.48\% & 31.06\% & 37.70\% &\\
    GPT-4 ICL examples  &   &   & 40.86\% & 45.84\% & 36.86\% & 41.19\% &\\
    Human examples  &   &   & 45.01\% & 41.82\% & 29.89\% & 38.91\% & \\
    Mini-GPT-4  &   &   & 38.68\% & 40.29\% & 30.51\% & 36.49\% & \\
    GPT-4  &   &   & 39.33\% & 41.86\% & 31.08\% & 37.42\% &\\
    Claude  &   &   & 51.89\% & 50.09\% & 37.27\% & 46.42\% & \\
    Ours  &   &   & 45.29\% & 50.06\% & 39.66\% & 45.00\% & \\
    \hline
    Ours - Claude  &   &   & \textcolor{red}{-6.60}\% & \textcolor{red}{-0.03}\% & \textcolor{ForestGreen}{+2.39}\% & \textcolor{red}{-1.42}\% &\\
    Ours - Avg of Others  &   &   & \textcolor{ForestGreen}{+3.07}\% & \textcolor{ForestGreen}{+5.99}\% & \textcolor{ForestGreen}{+6.89}\% & \textcolor{ForestGreen}{+5.31}\% &\\

    \hline
    Upper bound  & All Data & 18.0 & 52.88\% & 54.87\% & 41.87\% & 49.88\% &   \\
    skywork  &   &   & 43.66\% & 46.01\% & 36.12\% & 41.93\% & 0.101 \\
    CAR  &   &   & 41.87\% & 44.99\% & 35.00\% & 40.62\% & 0.133 \\
    IFD  &   &   & 52.88\% & 49.73\% & 38.19\% & 46.93\% & 0.386 \\
    perplexity  &   &   & 42.69\% & 48.50\% & 36.44\% & 42.54\% & 0.198 \\
    Ours  &   &   & 45.29\% & 50.06\% & 39.66\% & 45.00\% & 0.241 \\
    \hline
    Ours - Perplexity  &   &   & \textcolor{ForestGreen}{+2.59}\% & \textcolor{ForestGreen}{+1.56}\% & \textcolor{ForestGreen}{+3.22}\% & \textcolor{ForestGreen}{+2.46}\% & \textcolor{ForestGreen}{+0.043} \\
    \hline
    
  \end{tabular}}
\caption{Comparison of our method with other metrics or response generation methods on 6 subsets from the PlanBench dataset as introduced by \hyperref[paragraph:PlanBench extension]{PlanBench Extension}, introduced in Section 4.2. We compare our method against IFD \citep{li-etal-2024-quantity}, Skywork \citep{liu2024skyworkrewardbagtricksreward}, CAR \citep{xu2024strongermodelsstrongerteachers}, and Perplexity\citep{ren2024learn}.}  
  \label{tab:planbench}
\end{table*}

\subsection{Comparison with Alternative Response Selection Metrics}
\label{sec:metric_comparison}

Table~\ref{tab:icppl vs other metrics: average of 3 runs} reports results obtained with the same set-up as in Section~\ref{experiment_all_data}, but swapping the ranking metric.  
Across the full \hyperref[paragraph: Main-experiment corpus]{Main-experiment corpus}, \emph{self-aligned perplexity} achieves the best mean accuracy and the highest weighted Spearman correlation; standard perplexity is the closest baseline. \textbf{All tasks:} Using every training run, our metric surpasses standard perplexity by 0.33\% in accuracy and by 0.033 in weighted Spearman~$\rho$. \textbf{High-variance tasks (\boldmath$\mathrm{SD}>2\%$):} The margins widen to 0.80\% in accuracy and 0.032 in weighted~$\rho$. \textbf{Very-high-variance tasks (\boldmath$\mathrm{SD}>4\%$):} Gains further increase to 1.76\% in accuracy and 0.075 in weighted~$\rho$.
These results mirror the trend observed in Section \ref{experiment_all_data}: the larger the performance spread among candidate strategies, the more our metric outperforms conventional perplexity, underscoring its value for selecting high-quality training data.

\subsection{Performance Differences among Response Generation Strategies Can be Very Large \label{Performance Differences among Response Generation Strategies Can be Very Large}}

Candidate response-generation strategies can yield significantly different results depending on the task.
To illustrate this, we evaluate various strategies on the \hyperref[paragraph:PlanBench extension]{\textsc{PlanBench}} benchmark introduced in Section~\ref{datasets}, which is designed to be more difficult than standard instruction-following datasets due to its long-horizon, goal-conditioned reasoning requirements.
As shown in Table~\ref{tab:planbench} and Tabel~\ref{tab:ntrain_1000_lr_2e-05_seedaverage_of_seed_0,1_pan_bench}(Appendix), training outcomes vary significantly across methods, underscoring the importance of selecting an appropriate generation strategy. Our self-aligned perplexity metric improves accuracy by an average of $2.46\%$ over standard perplexity and $5.31\%$ over the mean performance of all strategies. The results further demonstrate that, as the optimal model varies across datasets and continues to shift as APIs evolve, model-aware selection metrics like self-aligned perplexity remain critical. 

As shown in Table~\ref{tab:planbench}, the average accuracy of IFD is slightly higher than that of our method. However, our method shows comparatively better performance on the main tasks, which cover a broader set of 17 tasks (Table~\ref{tab:icppl vs other metrics: average of 3 runs}). This suggests that our approach may offer more stable and reliable results when applied across a wider range of tasks.



\section{Ablation Study}

\subsection{Why Self-Aligned Perplexity Outperforms Traditional Perplexity}
\label{sec:EmpiricalStudyOfWhyDoesSelfAlignedPerplexityWorkBetterThanTraditionalPerplexity}

\begin{table}[t]
\centering
\resizebox{0.48\textwidth}{!}{
\begin{tabular}{l|l|c|c|c|c|c}
\hline
Target Response style & Model & Task  & \texttt{PPL} &  \texttt{S}$_{\text{sbs}}$\texttt{PPL}    & \texttt{S}$_{\text{cad}}$\texttt{PPL}  & \texttt{S}$_{\text{r}}$\texttt{PPL}  \\ \hline
Step by Step(sbs) &  Mistral7B & ECQA     & \textbf{4.476}  & \textbf{3.695} & 4.85    &  4.329    \\ 
GPT4 Answer Directly(cad)   &        &  & 5.551  & 4.116 & \textbf{4.768}    &   4.456  \\ 
Redundant(r)    &      & & 4.944 & \textcolor{red}{4.334} &  \textcolor{red}{5.615}   &   \textbf{4.326}   \\  
\hline
Step by Step(sbs) &  Mistral7B &     PIQA    &    \textbf{4.290}    &  \textbf{3.816}  & 5.968 & 4.028\\ 
GPT4 Answer Directly(cad)  &      &    & 6.277 & 4.053 &  \textbf{5.962} & 4.250\\ 
Redundant(r)   & &   &  4.547  & 3.919 & \textcolor{red}{6.724} &  \textbf{4.027} \\  
\hline
\end{tabular}}
\caption{Examples showing that in-context perplexity favors responses matching the style of the in-context example. \texttt{PPL} is standard perplexity; \texttt{S}$_{\text{sbs}}$\texttt{PPL}, \texttt{S}$_{\text{cad}}$\texttt{PPL}, and \texttt{S}$_{\text{r}}$\texttt{PPL} use step-by-step, GPT-4o Answer Directly, and redundant responses as context, respectively.}
\label{tab:self-aligned perplexity push perplexity to the right direction}
\end{table}

Traditional perplexity is sensitive to surface-level stylistic cues, so a low score does not necessarily mean the response “feels” familiar to the model.
We therefore anchor the metric on the model’s own zero-shot prediction: the closer a candidate lies to this anchor, the more familiar it should be.
Injecting that prediction as a single in-context example reshapes the probability distribution, yielding a self-aligned perplexity that more faithfully reflects the response’s true familiarity.

\paragraph{When perplexity fails.} 
\label{failure case of perplexity}
According to Table~\ref{tab:self-aligned perplexity push perplexity to the right direction}, 
On ECQA, a deliberately \textit{redundant} answer(see Appendix~\ref{appendix:redundant_prompt} for how we construct this dataset) scores 4.94 in raw perplexity, while the terser, higher-quality \textit{GPT-4-direct} answer scores 5.55 (Table~\ref{tab:self-aligned perplexity push perplexity to the right direction}).
A similar pattern appears in PIQA (4.55 vs.\ 6.28).
Thus, lower perplexity can sometimes reflect wordiness rather than genuine familiarity with the model’s preferred style.

\paragraph{How self-aligned perplexity helps.}
According to Table~\ref{tab:self-aligned perplexity push perplexity to the right direction}, adding a single in-context example can realign the perplexity scores. For the GPT-4–style response on ECQA, the raw perplexity (PPL) is 5.551, which is higher than the redundant-style response (4.944). After prepending an in-context example drawn from another GPT-4–style answer, the GPT-4 response’s perplexity drops to 4.768. In contrast, when the same example is added to the redundant and step-by-step responses, their perplexities increase from 4.944 and 4.476 to 5.615 and 4.850, respectively.

A similar effect occurs on PIQA: the GPT-4 response on Mistral has an initial perplexity of 6.277, higher than the redundant style (4.570). With a GPT-4 in-context example, its perplexity decreases to 5.962, while the redundant style’s perplexity rises to 6.724.

Across the two tasks, using the model’s own prediction as the in-context anchor consistently lowers the score for its native style by 0.6–1.5 points, restoring the correct ordering and yielding rankings that track downstream fine-tuning gains.


\begin{table}[t]
  \centering
  \small
  \resizebox{0.45\textwidth}{!}{%
    \begin{tabular}{l|l|c|c}
      \hline
      Weighted \textbf{Method} & \textbf{Model} & \textbf{Accuracy} & \textbf{Spearman's $\rho$} \\ \hline
      Ours             & Mistral7B & 0.583 & 0.317 \\
      TTT (lr=2e-5)    &          & 0.578  & 0.188 \\
      TTT (lr=2e-4)    &          & 0.579 & 0.557 \\ \hline
      Ours             & Llama3    &  0.637 & 0.288 \\
      TTT (lr=2e-5)    &           & 0.613 & 0.168 \\
      TTT (lr=2e-4)    &           & 0.627 & 0.398 \\ \hline
      Ours             & Qwen2.5   &  0.705 &  0.367 \\
      TTT (lr=2e-5)    &          & 0.702 & 0.065  \\
      TTT (lr=2e-4)    &           & 0.709 & 0.273 \\ \hline
      Ours             & Average   & 0.642 & 0.324\\
      TTT (lr=2e-5)    &           &  0.631 & 0.141 \\
      TTT (lr=2e-4)    &           & 0.638 &  0.409 \\ \hline
    \end{tabular}%
  }
  \caption{Ours (K=50, avg. of 3 subsets) vs.\ Train-then-Test (TTT) (K=100, 1 seed) on \hyperref[paragraph:Main-experiment corpus]{Main-experiment corpus}.}
  \label{tab:ours_vs_train_then_test}
\end{table}

\begin{table*}[t]
\centering
\small
\resizebox{0.78\textwidth}{!}{
\begin{tabular}{l|c|c|c|c|c|c|c|c|c}
\hline
 & 0--10 & 10--20 & 20--30 & 0--30 & 30--60 & 60--90 & 0--50 & 50--100 & 100--150 \\ 
\hline
Accuracy (\%)      & 63.90\% & 64.22\% &  63.93\% &  64.25\%  &  63.96\%& 64.22\% & 64.17\%  & 64.10\% &  64.22\%\\
Weighted $\rho$      &  0.289 &  0.363 &   0.290 &   0.327  & 0.286 &  0.269 &  0.317 &  0.288 &   0.367 \\
\hline
 & 0--100 & 100--200 & 200--300 & 0--200 & 0--300 & - & - & - & - \\ 
\hline
Accuracy (\%)      & 64.18\%  & 64.09\% &  64.08\%  & 64.13\%  & 64.13\%  & - & - & - & - \\
Weighted $\rho$    & 0.332  &  0.330 & 0.323  &  0.333  &   0.342 & - & - & - & - \\
\hline
\end{tabular}}
\caption{Performance on different subsets when ranking with self-aligned perplexity.  
An interval such as \emph{60–90} means starting at index 60 and using the next 30 instances (indices 60–89) for ranking calculation.}
\label{tab:stability_our_method}
\end{table*}




\subsection{Why Do Some Datasets Show Greater Variance in Training Outcomes? \label{Why Do Some Datasets Show Greater Variance in Training Outcomes?}}

\begin{figure}[t]
\centering
\includegraphics[width=0.80\columnwidth]{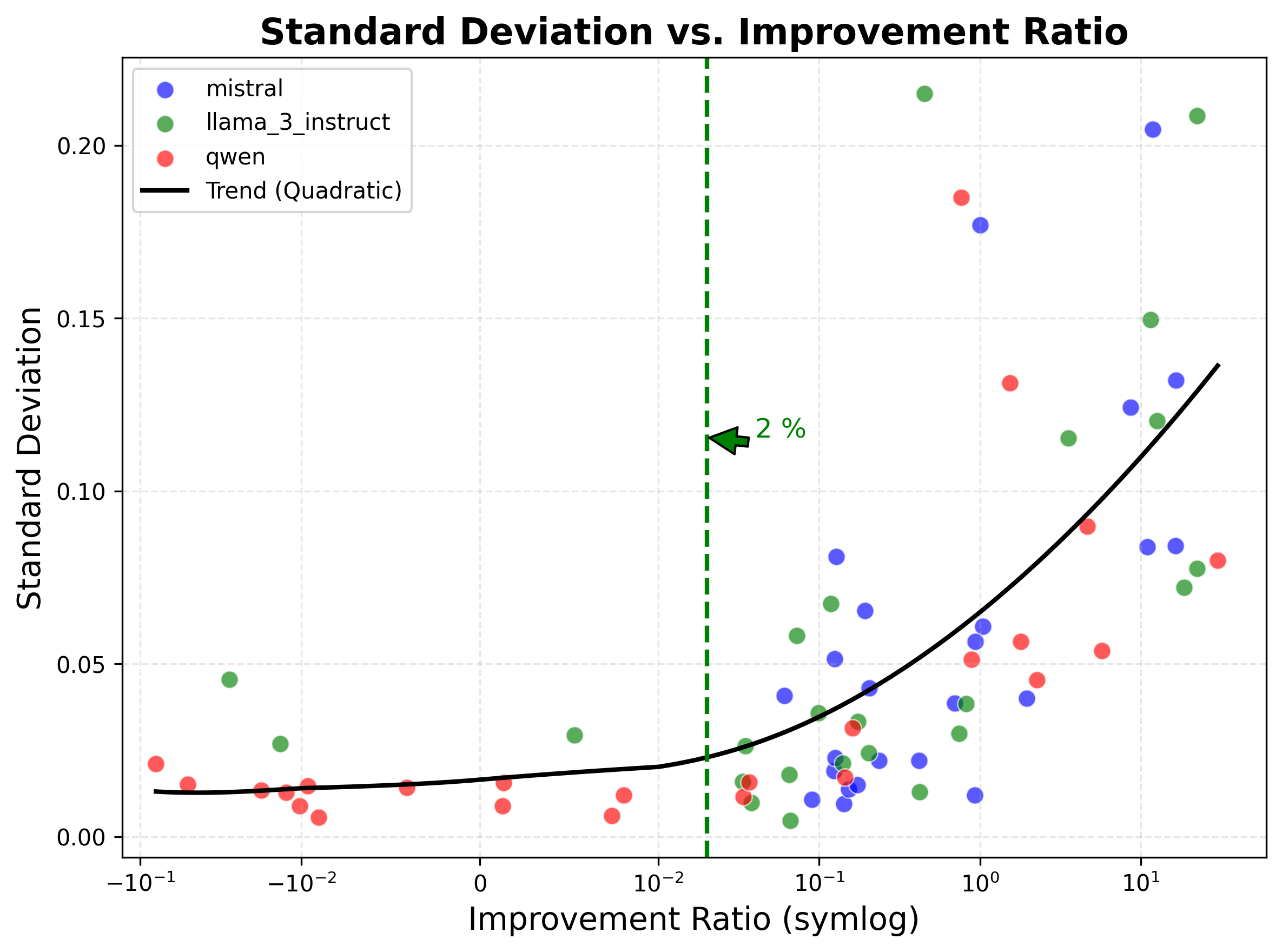}
\caption{When the improvement ratio is high, the standard deviation of training outcomes across different response‐generation strategies tends to be larger.}

\label{Diversity_vs_Improved_Percentage_2_plots}
\end{figure}

\begin{table*}[t!]
  \centering
  \resizebox{1.0\textwidth}{!}{%
  \begin{tabular}{l|c|c|c|c|c|c|c|c|c|c|c|c|c|c|c|c}
    \hline
    Model  & gsm8k & math algebra & mmlu & winogrande & piqa & agieval & squad & ecqa & boolq & arc challenge & mmlu pro law & drop & hellaswag & mmlu moral scenarios & math geometry & api bank \\ \hline
    qwen & 94.8\% / 88.9\% & 93.4\% / 91.6\% & 51.0\% / 50.9\% & 72.7\% / 72.1\% & 86.0\% / \textbf{87.9\%} & 57.2\% / 54.5\% & 67.3\% / \textbf{76.2\%} & 81.6\% / 79.3\% & 83.6\% / \textbf{86.5\%} & 89.3\% / 89.0\% & 32.1\% / 27.6\% & 0.7\% / \textbf{79.4\%} & 70.6\% / 70.6\% & 56.8\% / \textbf{63.3\%} & 51.9\% / \textbf{65.8\%} & 11.7\% / \textbf{43.3\%} \\
    mistral & 39.3\% / \textbf{62.5\%} & 15.4\% / \textbf{31.9\%} & 37.6\% / \textbf{42.2\%} & 44.9\% / \textbf{71.3\%} & 69.2\% / \textbf{86.9\%} & 33.5\% / \textbf{40.0\%} & 8.8\% / \textbf{73.2\%} & 57.6\% / \textbf{70.0\%} & 82.3\% / \textbf{86.7\%} & 73.5\% / 61.1\% & 19.4\% / \textbf{22.9\%} & 0.6\% / \textbf{74.6\%} & 39.7\% / \textbf{65.4\%} & 37.2\% / \textbf{71.3\%} & 5.1\% / \textbf{17.7\%} & 1.8\% / \textbf{51.0\%} \\
    llama 3 instruct & 85.8\% / 81.4\% & 58.1\% / 56.2\% & 43.0\% / \textbf{48.9\%} & 68.4\% / \textbf{69.5\%} & 81.1\% / \textbf{86.5\%} & 43.0\% / \textbf{43.5\%} & 24.5\% / \textbf{75.2\%} & 72.0\% / \textbf{72.3\%} & 78.1\% / \textbf{88.0\%} & 79.4\% / \textbf{80.1\%} & 25.9\% / \textbf{27.6\%} & 15.1\% / \textbf{79.6\%} & 56.8\% / \textbf{72.2\%} & 33.3\% / \textbf{67.7\%} & 18.3\% / \textbf{27.8\%} & 0.0\% / \textbf{49.4\%} \\
    \hline
  \end{tabular}}
  \caption{Zero-shot (left) vs. trained on GPT-4o direct-answer data (right) accuracy across tasks for each model. Win ratios of GPT-4o direct-answer training over zero-shot: qwen: 7/16 (43.8\%); mistral: 15/16 (93.8\%); llama 3 instruct: 14/16 (87.5\%).}
\label{tab:win_ratio_record}
\end{table*}

We observed a striking regularity across tasks: whenever enlarging the training set from 100 to 1 000 examples yields little or no accuracy gain, the choice of response-generation method matters equally little. Conversely, tasks that continue to improve with more data show pronounced performance gaps between generation strategies.

Let the \textbf{improvement ratio} be defined as $\text{Acc}_{1000} / \text{Acc}_{100} - 1$, representing the relative gain from increasing the training size ten-fold.
Figure~\ref{Diversity_vs_Improved_Percentage_2_plots} plots $\log(\text{improvement ratio})$ (x-axis) against the standard deviation of accuracies across generation methods (y-axis). A clear positive trend emerges: once the improvement ratio exceeds roughly 2\%, the variance among methods rises sharply; below this threshold, it is nearly zero.  We plot Figure~\ref{Diversity_vs_Improved_Percentage_2_plots} using training results from all tasks in the Main-experiment corpus and PlanBench.

The results suggest that divergence across generation strategies is greatest exactly when the dataset still offers headroom for improvement. On such high-variance tasks, selecting the right response-generation method is critical, underscoring the value of our self-aligned perplexity criterion.


Figure \ref{Diversity_vs_Improved_Percentage_2_plots} shows that many red data points lie to the left of the green 2\% threshold line, indicating that training with data generated by different response-generation strategies produces negligible variation in training outcomes on Qwen. We hypothesize that this low-variance pattern arises because the training set contributes little information beyond Qwen’s existing capabilities. This also explains why our method improves accuracy by only 0.45\% over the average of all response-generation strategies on Qwen when evaluated on the full dataset.
In Table \ref{tab:win_ratio_record}, we compare zero-shot accuracy with the accuracy of the three models after being trained on GPT-4o direct-answer data for each task. As shown, only 43.8\% of tasks improve after training, suggesting that Qwen already possesses strong prior knowledge of standard benchmarks for most tasks. In contrast, when trained on data with STD > 4\%, Qwen achieves a 3.38\% accuracy improvement over the average of all other methods.
On PlanBench, as demonstrated in Section \ref{Performance Differences among Response Generation Strategies Can be Very Large} and Table \ref{tab:planbench}, our method yields substantial accuracy gains across the six PlanBench subtasks with Qwen: +3.22\% over perplexity and +6.89\% over the average of all strategies.

\subsection{The effectiveness of Correctness Filter}\label{sec:effectiveness of correctness filter}
Correctness filter is helpful mainly on the tasks that the model is very unfamiliar with.

Removing the correctness filter has negligible impact on the 17 main tasks, as shown in the Table~\ref{tab:ours_vs_wofilter}. The table  shows the accuracy and weighted Spearman correlation deltas between the filtered and unfiltered versions of our method (Ours \- Ours-w/o-Filter); Overall, skipping the filter—i.e., using the model’s raw predictions—only minimally reduces performance in terms of average accuracy and weighted Spearman correlation.

However, removing the correctness filter results in a  performance drop on PlanBench, as shown in the table~\ref{tab:ours_vs_wofilter_single}.

\subsection{Our Method vs. Train-Then-Select}
One natural (but computationally expensive) approach to select the optimal response generation strategy is to adopt a Train-Then-Select (TTS) procedure. In this way, we first generate a small dataset (e.g., 100 samples) using each candidate strategy. For each dataset, we train the target model and evaluate its performance. We then rank the strategies based on the results and choose the best-performing one to generate the remainder of the dataset. 

When evaluating TTS, we train the target model on 100 samples under two settings:  
\textbf{1) Standard Training:} A learning rate of 2e-5 for 20 epochs (matching our main setup). The performance accuracies for each strategy under this setting is in the Table~\ref{tab:ntrain_100_lr_2e-05_seed0}.
\textbf{2) Intense Training:} A learning rate of 2e-4 for 40 epochs. The performance accuracies for each strategy under this setting is in Table~\ref{tab:ntrain_100_lr_0.0002_seed0}.

After ranking the strategies using TTS, we compare their performance with ours. In Table~\ref{tab:ours_vs_train_then_test}, despite using less data and requiring no training, validation, or testing computations for strategy selection, our method achieves better average accuracy and comparable weighted Spearman correlation.

\subsection{Stability of Our Method}

As shown in Table~\ref{tab:stability_our_method}, accuracy generally improves as the subset size grows, and the overall performance is consistent across ranges. Small subsets sometimes degrade accuracy (values highlighted in red); thus, we recommend using at least 30 samples or even 50 samples for the best performance. Specifically, “0–10”, “10–20”, and “20–30” denote the first, second, and third batches of ten training examples, respectively, while “0–50” and “0–100” correspond to the first 50 and 100 examples. When the subset size reaches 50 or more, average accuracy stabilises. The weighted Spearman correlation ($\rho$) also increases with larger subsets, but the gains taper off once the subset size exceeds 50.

\section{Conclusions}

In this paper, we present a novel and scalable approach for selecting the optimal response-generation strategy to train large language models. We introduce a new metric, self-aligned perplexity, which more effectively evaluates the alignment between a target model and its response options compared to traditional perplexity. We demonstrate that choosing the optimal generation strategy based on self-aligned perplexity leads to substantial improvements in model performance, particularly on tasks with high performance variance. We hope our work will inspire researchers who use perplexity as a downstream metric or who wish to build the most effective instruction tuning datasets.

\section{Limitations}


While recent open-source ``thinking'' models(such as DeepSeek-R1-Distill-Qwen-7B) support long chain-of-thought reasoning, our evaluations focus on chat models. We used meta prompts to elicit reasoning steps but did not test on ``thinking'' models. We believe our style-alignment approach still applies, though further validation on smaller or differently pretrained models is needed. We leave broader scaling studies and extensions to other model families for future work.

\bibliography{custom}

\begin{thebibliography}{45}
\providecommand{\natexlab}[1]{#1}

\bibitem[{Aggarwal et~al.(2021)Aggarwal, Mandowara, Agrawal, Khandelwal, Singla, and Garg}]{aggarwal2021explanations}
Shourya Aggarwal, Divyanshu Mandowara, Vishwajeet Agrawal, Dinesh Khandelwal, Parag Singla, and Dinesh Garg. 2021.
\newblock Explanations for commonsenseqa: New dataset and models.
\newblock In \emph{Proceedings of the 59th Annual Meeting of the Association for Computational Linguistics and the 11th International Joint Conference on Natural Language Processing (Volume 1: Long Papers)}, pages 3050--3065.

\bibitem[{Anthropic(2023)}]{anthropic_claude_api}
Anthropic. 2023.
\newblock Claude 3.5 api.
\newblock \url{https://docs.anthropic.com/claude}.
\newblock Accessed: Month Day, Year.

\bibitem[{Austin et~al.(2021)Austin, Odena, Nye, Bosma, Michalewski, Dohan, Jiang, Cai, Terry, Le et~al.}]{austin2021program}
Jacob Austin, Augustus Odena, Maxwell Nye, Maarten Bosma, Henryk Michalewski, David Dohan, Ellen Jiang, Carrie Cai, Michael Terry, Quoc Le, and 1 others. 2021.
\newblock Program synthesis with large language models.
\newblock \emph{arXiv preprint arXiv:2108.07732}.

\bibitem[{Bisk et~al.(2020)Bisk, Zellers, Gao, Choi et~al.}]{bisk2020piqa}
Yonatan Bisk, Rowan Zellers, Jianfeng Gao, Yejin Choi, and 1 others. 2020.
\newblock Piqa: Reasoning about physical commonsense in natural language.
\newblock In \emph{Proceedings of the AAAI conference on artificial intelligence}, volume~34, pages 7432--7439.

\bibitem[{Clark et~al.(2019)Clark, Lee, Chang, Kwiatkowski, Collins, and Toutanova}]{clark2019boolq}
Christopher Clark, Kenton Lee, Ming-Wei Chang, Tom Kwiatkowski, Michael Collins, and Kristina Toutanova. 2019.
\newblock Boolq: Exploring the surprising difficulty of natural yes/no questions.
\newblock \emph{arXiv preprint arXiv:1905.10044}.

\bibitem[{Clark et~al.(2018)Clark, Cowhey, Etzioni, Khot, Sabharwal, Schoenick, and Tafjord}]{clark2018think}
Peter Clark, Isaac Cowhey, Oren Etzioni, Tushar Khot, Ashish Sabharwal, Carissa Schoenick, and Oyvind Tafjord. 2018.
\newblock Think you have solved question answering? try arc, the ai2 reasoning challenge.
\newblock \emph{arXiv preprint arXiv:1803.05457}.

\bibitem[{Cobbe et~al.(2021)Cobbe, Kosaraju, Bavarian, Chen, Jun, Kaiser, Plappert, Tworek, Hilton, Nakano et~al.}]{cobbe2021training}
Karl Cobbe, Vineet Kosaraju, Mohammad Bavarian, Mark Chen, Heewoo Jun, Lukasz Kaiser, Matthias Plappert, Jerry Tworek, Jacob Hilton, Reiichiro Nakano, and 1 others. 2021.
\newblock Training verifiers to solve math word problems.
\newblock \emph{arXiv preprint arXiv:2110.14168}.

\bibitem[{De~la Rosa et~al.(2022)De~la Rosa, Ponferrada, Villegas, Salas, Romero, and Grandury}]{de2022bertin}
Javier De~la Rosa, Eduardo~G Ponferrada, Paulo Villegas, Pablo Gonzalez de~Prado Salas, Manu Romero, and Mar{\i}a Grandury. 2022.
\newblock Bertin: Efficient pre-training of a spanish language model using perplexity sampling.
\newblock \emph{arXiv preprint arXiv:2207.06814}.

\bibitem[{Dua et~al.(2019)Dua, Wang, Dasigi, Stanovsky, Singh, and Gardner}]{dua2019dropreadingcomprehensionbenchmark}
Dheeru Dua, Yizhong Wang, Pradeep Dasigi, Gabriel Stanovsky, Sameer Singh, and Matt Gardner. 2019.
\newblock \href {https://arxiv.org/abs/1903.00161} {Drop: A reading comprehension benchmark requiring discrete reasoning over paragraphs}.
\newblock \emph{Preprint}, arXiv:1903.00161.

\bibitem[{Dubey and Abhinav~Jauhri(2024)}]{dubey2024llama3herdmodels}
Abhimanyu Dubey and etc. Abhinav~Jauhri. 2024.
\newblock \href {https://arxiv.org/abs/2407.21783} {The llama 3 herd of models}.
\newblock \emph{Preprint}, arXiv:2407.21783.

\bibitem[{Fu et~al.(2023)Fu, Peng, Ou, Sabharwal, and Khot}]{fu2023specializing}
Yao Fu, Hao Peng, Litu Ou, Ashish Sabharwal, and Tushar Khot. 2023.
\newblock Specializing smaller language models towards multi-step reasoning.
\newblock In \emph{International Conference on Machine Learning}, pages 10421--10430. PMLR.

\bibitem[{Gonen et~al.(2022)Gonen, Iyer, Blevins, Smith, and Zettlemoyer}]{gonen2022demystifying}
Hila Gonen, Srini Iyer, Terra Blevins, Noah~A Smith, and Luke Zettlemoyer. 2022.
\newblock Demystifying prompts in language models via perplexity estimation.
\newblock \emph{arXiv preprint arXiv:2212.04037}.

\bibitem[{Hendrycks et~al.(2020)Hendrycks, Burns, Basart, Zou, Mazeika, Song, and Steinhardt}]{hendrycks2020measuring}
Dan Hendrycks, Collin Burns, Steven Basart, Andy Zou, Mantas Mazeika, Dawn Song, and Jacob Steinhardt. 2020.
\newblock Measuring massive multitask language understanding.
\newblock \emph{arXiv preprint arXiv:2009.03300}.

\bibitem[{Hendrycks et~al.(2021)Hendrycks, Burns, Kadavath, Arora, Basart, Tang, Song, and Steinhardt}]{hendrycks2021measuring}
Dan Hendrycks, Collin Burns, Saurav Kadavath, Akul Arora, Steven Basart, Eric Tang, Dawn Song, and Jacob Steinhardt. 2021.
\newblock Measuring mathematical problem solving with the math dataset.
\newblock \emph{arXiv preprint arXiv:2103.03874}.

\bibitem[{Ho et~al.(2022)Ho, Schmid, and Yun}]{ho2022large}
Namgyu Ho, Laura Schmid, and Se-Young Yun. 2022.
\newblock Large language models are reasoning teachers.
\newblock \emph{arXiv preprint arXiv:2212.10071}.

\bibitem[{Hsieh et~al.(2023)Hsieh, Li, Yeh, Nakhost, Fujii, Ratner, Krishna, Lee, and Pfister}]{hsieh2023distilling}
Cheng-Yu Hsieh, Chun-Liang Li, Chih-Kuan Yeh, Hootan Nakhost, Yasuhisa Fujii, Alexander Ratner, Ranjay Krishna, Chen-Yu Lee, and Tomas Pfister. 2023.
\newblock \href {https://arxiv.org/abs/2305.02301} {Distilling step-by-step! outperforming larger language models with less training data and smaller model sizes}.
\newblock \emph{Preprint}, arXiv:2305.02301.

\bibitem[{Hu et~al.(2020)Hu, Gauthier, Qian, Wilcox, and Levy}]{hu2020systematic}
Jennifer Hu, Jon Gauthier, Peng Qian, Ethan Wilcox, and Roger~P Levy. 2020.
\newblock A systematic assessment of syntactic generalization in neural language models.
\newblock \emph{arXiv preprint arXiv:2005.03692}.

\bibitem[{Jiang et~al.(2023)Jiang, Sablayrolles, Mensch, Bamford, Chaplot, Casas, Bressand, Lengyel, Lample, Saulnier et~al.}]{jiang2023mistral}
Albert~Q Jiang, Alexandre Sablayrolles, Arthur Mensch, Chris Bamford, Devendra~Singh Chaplot, Diego de~las Casas, Florian Bressand, Gianna Lengyel, Guillaume Lample, Lucile Saulnier, and 1 others. 2023.
\newblock Mistral 7b.
\newblock \emph{arXiv preprint arXiv:2310.06825}.

\bibitem[{Kang et~al.(2023)Kang, Lee, Baek, Kawaguchi, and Hwang}]{kang2023knowledgeaugmented}
Minki Kang, Seanie Lee, Jinheon Baek, Kenji Kawaguchi, and Sung~Ju Hwang. 2023.
\newblock \href {https://arxiv.org/abs/2305.18395} {Knowledge-augmented reasoning distillation for small language models in knowledge-intensive tasks}.
\newblock \emph{Preprint}, arXiv:2305.18395.

\bibitem[{Kim et~al.(2024)Kim, Suk, Yue, Viswanathan, Lee, Wang, Gashteovski, Lawrence, Welleck, and Neubig}]{kim2024evaluatinglanguagemodelssynthetic}
Seungone Kim, Juyoung Suk, Xiang Yue, Vijay Viswanathan, Seongyun Lee, Yizhong Wang, Kiril Gashteovski, Carolin Lawrence, Sean Welleck, and Graham Neubig. 2024.
\newblock \href {https://arxiv.org/abs/2412.03679} {Evaluating language models as synthetic data generators}.
\newblock \emph{Preprint}, arXiv:2412.03679.

\bibitem[{Kocmi and Bojar(2017)}]{kocmi2017curriculum}
Tom Kocmi and Ondrej Bojar. 2017.
\newblock Curriculum learning and minibatch bucketing in neural machine translation.
\newblock \emph{arXiv preprint arXiv:1707.09533}.

\bibitem[{Li et~al.(2024)Li, Zhang, Li, Chen, Chen, Cheng, Wang, Zhou, and Xiao}]{li-etal-2024-quantity}
Ming Li, Yong Zhang, Zhitao Li, Jiuhai Chen, Lichang Chen, Ning Cheng, Jianzong Wang, Tianyi Zhou, and Jing Xiao. 2024.
\newblock \href {https://doi.org/10.18653/v1/2024.naacl-long.421} {From quantity to quality: Boosting {LLM} performance with self-guided data selection for instruction tuning}.
\newblock In \emph{Proceedings of the 2024 Conference of the North American Chapter of the Association for Computational Linguistics: Human Language Technologies (Volume 1: Long Papers)}, pages 7602--7635, Mexico City, Mexico. Association for Computational Linguistics.

\bibitem[{Li et~al.(2023)Li, Zhao, Yu, Song, Li, Yu, Li, Huang, and Li}]{li2023api}
Minghao Li, Yingxiu Zhao, Bowen Yu, Feifan Song, Hangyu Li, Haiyang Yu, Zhoujun Li, Fei Huang, and Yongbin Li. 2023.
\newblock Api-bank: A comprehensive benchmark for tool-augmented llms.
\newblock \emph{arXiv preprint arXiv:2304.08244}.

\bibitem[{Li et~al.(2022)Li, Chen, Shen, Chen, Zhang, Li, Wang, Qian, Peng, Mao, Chen, and Yan}]{li2022explanations}
Shiyang Li, Jianshu Chen, Yelong Shen, Zhiyu Chen, Xinlu Zhang, Zekun Li, Hong Wang, Jing Qian, Baolin Peng, Yi~Mao, Wenhu Chen, and Xifeng Yan. 2022.
\newblock \href {https://arxiv.org/abs/2210.06726} {Explanations from large language models make small reasoners better}.
\newblock \emph{Preprint}, arXiv:2210.06726.

\bibitem[{Liu et~al.(2024)Liu, Zeng, Liu, Yan, He, Wang, Yan, Liu, and Zhou}]{liu2024skyworkrewardbagtricksreward}
Chris~Yuhao Liu, Liang Zeng, Jiacai Liu, Rui Yan, Jujie He, Chaojie Wang, Shuicheng Yan, Yang Liu, and Yahui Zhou. 2024.
\newblock \href {https://arxiv.org/abs/2410.18451} {Skywork-reward: Bag of tricks for reward modeling in llms}.
\newblock \emph{Preprint}, arXiv:2410.18451.

\bibitem[{Luo et~al.(2023)Luo, Xu, Zhao, Sun, Geng, Hu, Tao, Ma, Lin, and Jiang}]{luo2023wizardcoderempoweringcodelarge}
Ziyang Luo, Can Xu, Pu~Zhao, Qingfeng Sun, Xiubo Geng, Wenxiang Hu, Chongyang Tao, Jing Ma, Qingwei Lin, and Daxin Jiang. 2023.
\newblock \href {https://arxiv.org/abs/2306.08568} {Wizardcoder: Empowering code large language models with evol-instruct}.
\newblock \emph{Preprint}, arXiv:2306.08568.

\bibitem[{Magister et~al.(2023)Magister, Mallinson, Adamek, Malmi, and Severyn}]{magister-etal-2023-teaching}
Lucie~Charlotte Magister, Jonathan Mallinson, Jakub Adamek, Eric Malmi, and Aliaksei Severyn. 2023.
\newblock \href {https://doi.org/10.18653/v1/2023.acl-short.151} {Teaching small language models to reason}.
\newblock In \emph{Proceedings of the 61st Annual Meeting of the Association for Computational Linguistics (Volume 2: Short Papers)}, pages 1773--1781, Toronto, Canada. Association for Computational Linguistics.

\bibitem[{Mekala et~al.(2024)Mekala, Nguyen, and Shang}]{mekala2024smaller}
Dheeraj Mekala, Alex Nguyen, and Jingbo Shang. 2024.
\newblock Smaller language models are capable of selecting instruction-tuning training data for larger language models.
\newblock \emph{arXiv preprint arXiv:2402.10430}.

\bibitem[{OpenAI(2023)}]{openai_gpt4_api}
OpenAI. 2023.
\newblock Gpt-4 api.
\newblock \url{https://platform.openai.com/docs/models/gpt-4}.
\newblock Accessed: Month Day, Year.

\bibitem[{Platanios et~al.(2019)Platanios, Stretcu, Neubig, Poczos, and Mitchell}]{platanios2019competence}
Emmanouil~Antonios Platanios, Otilia Stretcu, Graham Neubig, Barnabas Poczos, and Tom~M Mitchell. 2019.
\newblock Competence-based curriculum learning for neural machine translation.
\newblock \emph{arXiv preprint arXiv:1903.09848}.

\bibitem[{Qwen et~al.(2025)Qwen, :, Yang, Yang, Zhang, Hui, Zheng, Yu, Li, Liu, Huang, Wei, Lin, Yang, Tu, Zhang, Yang, Yang, Zhou, Lin, Dang, Lu, Bao, Yang, Yu, Li, Xue, Zhang, Zhu, Men, Lin, Li, Tang, Xia, Ren, Ren, Fan, Su, Zhang, Wan, Liu, Cui, Zhang, and Qiu}]{qwen2025qwen25technicalreport}
Qwen, :, An~Yang, Baosong Yang, Beichen Zhang, Binyuan Hui, Bo~Zheng, Bowen Yu, Chengyuan Li, Dayiheng Liu, Fei Huang, Haoran Wei, Huan Lin, Jian Yang, Jianhong Tu, Jianwei Zhang, Jianxin Yang, Jiaxi Yang, Jingren Zhou, and 25 others. 2025.
\newblock \href {https://arxiv.org/abs/2412.15115} {Qwen2.5 technical report}.
\newblock \emph{Preprint}, arXiv:2412.15115.

\bibitem[{Rajpurkar et~al.(2016)Rajpurkar, Zhang, Lopyrev, and Liang}]{rajpurkar-etal-2016-squad}
Pranav Rajpurkar, Jian Zhang, Konstantin Lopyrev, and Percy Liang. 2016.
\newblock \href {https://doi.org/10.18653/v1/D16-1264} {{SQ}u{AD}: 100,000+ questions for machine comprehension of text}.
\newblock In \emph{Proceedings of the 2016 Conference on Empirical Methods in Natural Language Processing}, pages 2383--2392, Austin, Texas. Association for Computational Linguistics.

\bibitem[{Ranaldi and Freitas(2024)}]{ranaldi-freitas-2024-aligning}
Leonardo Ranaldi and Andre Freitas. 2024.
\newblock \href {https://aclanthology.org/2024.eacl-long.109} {Aligning large and small language models via chain-of-thought reasoning}.
\newblock In \emph{Proceedings of the 18th Conference of the European Chapter of the Association for Computational Linguistics (Volume 1: Long Papers)}, pages 1812--1827, St. Julian{'}s, Malta. Association for Computational Linguistics.

\bibitem[{Ren et~al.(2024)Ren, Wu, and Liu}]{ren2024learn}
Xuan Ren, Biao Wu, and Lingqiao Liu. 2024.
\newblock \href {https://doi.org/10.18653/v1/2024.emnlp-main.571} {{I} learn better if you speak my language: Understanding the superior performance of fine-tuning large language models with {LLM}-generated responses}.
\newblock In \emph{Proceedings of the 2024 Conference on Empirical Methods in Natural Language Processing}, pages 10225--10245, Miami, Florida, USA. Association for Computational Linguistics.

\bibitem[{Sakaguchi et~al.(2021)Sakaguchi, Bras, Bhagavatula, and Choi}]{sakaguchi2021winogrande}
Keisuke Sakaguchi, Ronan~Le Bras, Chandra Bhagavatula, and Yejin Choi. 2021.
\newblock Winogrande: An adversarial winograd schema challenge at scale.
\newblock \emph{Communications of the ACM}, 64(9):99--106.

\bibitem[{Trinh et~al.(2024)Trinh, Wu, Le, He, and Luong}]{trinh2024solving}
Trieu~H Trinh, Yuhuai Wu, Quoc~V Le, He~He, and Thang Luong. 2024.
\newblock Solving olympiad geometry without human demonstrations.
\newblock \emph{Nature}, 625(7995):476--482.

\bibitem[{Valmeekam et~al.(2023)Valmeekam, Marquez, Olmo, Sreedharan, and Kambhampati}]{valmeekam2023planbench}
Karthik Valmeekam, Matthew Marquez, Alberto Olmo, Sarath Sreedharan, and Subbarao Kambhampati. 2023.
\newblock Planbench: An extensible benchmark for evaluating large language models on planning and reasoning about change.
\newblock \emph{Advances in Neural Information Processing Systems}, 36:38975--38987.

\bibitem[{Wang et~al.(2024)Wang, Ma, Zhang, Ni, Chandra, Guo, Ren, Arulraj, He, Jiang et~al.}]{wang2024mmlu}
Yubo Wang, Xueguang Ma, Ge~Zhang, Yuansheng Ni, Abhranil Chandra, Shiguang Guo, Weiming Ren, Aaran Arulraj, Xuan He, Ziyan Jiang, and 1 others. 2024.
\newblock Mmlu-pro: A more robust and challenging multi-task language understanding benchmark.
\newblock \emph{arXiv preprint arXiv:2406.01574}.

\bibitem[{Xu et~al.(2023)Xu, Sun, Zheng, Geng, Zhao, Feng, Tao, and Jiang}]{xu2023wizardlm}
Can Xu, Qingfeng Sun, Kai Zheng, Xiubo Geng, Pu~Zhao, Jiazhan Feng, Chongyang Tao, and Daxin Jiang. 2023.
\newblock \href {https://arxiv.org/abs/2304.12244} {Wizardlm: Empowering large language models to follow complex instructions}.
\newblock \emph{Preprint}, arXiv:2304.12244.

\bibitem[{Xu et~al.(2024)Xu, Jiang, Niu, Lin, and Poovendran}]{xu2024strongermodelsstrongerteachers}
Zhangchen Xu, Fengqing Jiang, Luyao Niu, Bill~Yuchen Lin, and Radha Poovendran. 2024.
\newblock \href {https://arxiv.org/abs/2411.07133} {Stronger models are not stronger teachers for instruction tuning}.
\newblock \emph{Preprint}, arXiv:2411.07133.

\bibitem[{Xu and Sheng(2024)}]{xu2024detecting}
Zhenyu Xu and Victor~S Sheng. 2024.
\newblock Detecting ai-generated code assignments using perplexity of large language models.
\newblock In \emph{Proceedings of the AAAI Conference on Artificial Intelligence}, volume~38, pages 23155--23162.

\bibitem[{Yang et~al.(2024)Yang, Pang, Feng, Wang, Chen, Zhu, and Liu}]{yang-etal-2024-self}
Zhaorui Yang, Tianyu Pang, Haozhe Feng, Han Wang, Wei Chen, Minfeng Zhu, and Qian Liu. 2024.
\newblock \href {https://doi.org/10.18653/v1/2024.acl-long.58} {Self-distillation bridges distribution gap in language model fine-tuning}.
\newblock In \emph{Proceedings of the 62nd Annual Meeting of the Association for Computational Linguistics (Volume 1: Long Papers)}, pages 1028--1043, Bangkok, Thailand. Association for Computational Linguistics.

\bibitem[{Zellers et~al.(2019)Zellers, Holtzman, Bisk, Farhadi, and Choi}]{zellers-etal-2019-hellaswag}
Rowan Zellers, Ari Holtzman, Yonatan Bisk, Ali Farhadi, and Yejin Choi. 2019.
\newblock \href {https://doi.org/10.18653/v1/P19-1472} {{H}ella{S}wag: Can a machine really finish your sentence?}
\newblock In \emph{Proceedings of the 57th Annual Meeting of the Association for Computational Linguistics}, pages 4791--4800, Florence, Italy. Association for Computational Linguistics.

\bibitem[{Zhang et~al.(2024)Zhang, Wang, Ao, and He}]{zhang2024distillation}
Hanyu Zhang, Xiting Wang, Xiang Ao, and Qing He. 2024.
\newblock Distillation with explanations from large language models.
\newblock In \emph{Proceedings of the 2024 Joint International Conference on Computational Linguistics, Language Resources and Evaluation (LREC-COLING 2024)}, pages 5018--5028.

\bibitem[{Zhong et~al.(2023)Zhong, Cui, Guo, Liang, Lu, Wang, Saied, Chen, and Duan}]{zhong2023agieval}
Wanjun Zhong, Ruixiang Cui, Yiduo Guo, Yaobo Liang, Shuai Lu, Yanlin Wang, Amin Saied, Weizhu Chen, and Nan Duan. 2023.
\newblock Agieval: A human-centric benchmark for evaluating foundation models.
\newblock \emph{arXiv preprint arXiv:2304.06364}.

\end{thebibliography}

\appendix

\section{Appendix}

\subsection{Ground Truth vs. Synthetic Data}
\label{sec:gt_vs_synthetic_data}
As shown in Table~\ref{tab:ntrain_1000_lr_2e-05_seedaverage_of_seed_0,1,2} (Appendix), when ground truth is provided in natural language (e.g., GSM8K, MATH, ECQA, MBPP), training on ground truth is less effective than on synthetic data. This is because LLMs are more familiar with LLM-generated data, as demonstrated by~\citet{ren2024learn}. However, when the ground truth is written as a gold label without a CoT inference process, training on the gold label can sometimes outperform training on CoT synthetic data within the same domain. However, in Table~\ref{table:cross_domain_performance_recording}~(Appendix), training on gold labels harms cross-domain performance more than training on synthetic data. Besides, in real-life scenarios, training on natural language data is crucial, as users expect to see the rationale behind the final prediction made by LLMs.

\subsection{Can we get performance gain if we simply put all of the response variants together?}

\begin{table*}[t]
\centering
\begin{tabular}{l|l|c|c|c}
\hline
\textbf{Method} & \textbf{Model} & \textbf{DROP} & \textbf{Hellaswag} &\textbf{API-Bank}  \\ \hline
Best $n_{\text{train}} = 1000$ & Mistral7B &0.743 & 0.675 & 0.559\\ 
Avg $n_{\text{train}} = 1000$ &  & 0.726 & 0.646 & 0.446 \\ 
Total $n_{\text{train}} = 6000$ & &0.740 & 0.738 & 0.555 \\ 
Mixture of good $n_{\text{train}} = 3000$ & &0.770 &  0.731& 0.555\\ 
Mixture of good $n_{\text{train}} = 1000$ & & 0.744& 0.686&0.535 \\
Average of all $n_{\text{train}} = 1000$ & & 0.711&0.686 &  0.433\\  
\hline
Best $n_{\text{train}} = 1000$& Llama3 &0.805  & 0.718 & 0.547\\
Avg $n_{\text{train}} = 1000$ &  & 0.778& 0.711 &0.392 \\ 
Total $n_{\text{train}} = 6000$ &  & 0.810 &0.738 & 0.490 \\ 
Mixture of good $n_{\text{train}} = 3000$ & &0.812 & 0.745 & 0.527\\ 
Mixture of good $n_{\text{train}} = 1000$ & & 0.804& 0.728& 0.490 \\  
Average of all $n_{\text{train}} = 1000$ & & 0.771 & 0.705& 0.457\\ 
\hline

\hline
Best $n_{\text{train}} = 1000$& Qwen2.5 & 0.814  & 0.739 & 0.461\\  
Avg $n_{\text{train}} = 1000$ &  &0.804 & 0.719 & 0.413 \\ 
Total $n_{\text{train}} = 6000$ &  & 0.798 & 0.748& 0.584\\ 
Mixture of good $n_{\text{train}} = 3000$ & & 0.824& 0.738 & 0.584\\ 
Mixture of good $n_{\text{train}} = 1000$ & & 0.818& 0.742& 0.490 \\ 
Average of all $n_{\text{train}} = 1000$ & &  0.778& 0.712& 0.412\\  
\hline

\end{tabular}
\caption{\textbf{Best} represents the best data generation strategy for the task with the target model. \textbf{Total} combines all strategies, yielding $n_{\text{train}} = 6000$. \textbf{Mixture of good} ($n_{\text{train}} = 3000$) includes the top three strategies with 1000 samples each, while \textbf{Mixture of good} ($n_{\text{train}} = 1000$) has about 333 samples per strategy.}
\label{tab:best_data_vs_all_data}
\end{table*}

Selecting the optimal data generation strategy remains essential, even when resources or funding are unlimited. As shown in Table~\ref{tab:best_data_vs_all_data}, simply combining six types of synthetic data (Total $n_{\text{train}} = 6000$) does not guarantee a performance gain over selecting the best synthetic training data. For example, after training the Llama3 model on API-Bank using all six types of synthetic data, the evaluation accuracy is only 49\%, much lower than when selecting the Claude Answer Directly data (54.7\%). Indeed, according to Table~\ref{tab:best_data_vs_all_data}, if we combine the mixture of the top three data generation strategies (Mixture of good $n_{\text{train}} = 3000$), the performance is almost always better than if we simply combine all of the data together (Total $n_{\text{train}} = 6000$). This underscores the importance of selecting data generation strategies, even if we can afford large-scale synthetic data generation and training.

\subsection{\textbf{The Impact of Accuracy of the Synthetic Data on Training Outcomes}}

In our experiment, we aim to ensure the correctness of generated answers by validating them against ground truth answers. Our research seeks to identify the best strategy for generating the optimal version of an answer. In other words, we can adjust data generation strategies to ensure correctness.

In our experiments, we use ground truth answers to guide the generated answers for nearly all datasets, with the only exceptions being mathematical problems. This follows the setting of the paper to maintain consistency with previous work \citep{ren2024learn}. This approach might be acceptable since closed-source APIs tend to generate accurate answers. For GSM8K and Math Algebra, GPT-4o, Claude, and MiniGPT-4o achieve accuracies of 90\% or above.

To evaluate the impact of accuracy on training outcomes, we conducted the following experiment. As shown in Table~\ref{tab:impact_accuracy}, we tested three approaches: training on the full dataset, using only correct predictions, and replacing incorrect predictions with rewritten ground truth. These approaches showed less than a 2\% improvement overall. Note that in this experiment, GPT-4 refers to the gpt-4-1106-preview API, rather than the gpt-4o-2024-08-06 API, which was used in all other experiments in the paper. The mathematical capabilities of GPT-4o, GPT-4-Mini, and Claude are similar on Math Algebra tasks. Therefore, we used the gpt-4-1106-preview API, which has a weaker ability to solve Math Algebra problems. The benifit of using it is that it makes more mistakes on GSM8K so that we can better evaluate the influence of accuracy. We used this API once to generate the data and train the model from there.

According to the table, the overall benefit of replacing incorrect examples with rewritten ground truth or removing incorrect examples has minimal impact on the overall training outcomes.

\subsection{Weighted Spearman’s Rank Correlation Coefficient}
\label{sec:appendix_weighted_spearman}

Spearman’s rank correlation ($\rho$) measures how well two orderings agree, ignoring absolute values.  
Because some of our \{model,\,dataset\} pairs exhibit far larger performance gaps among response-generation strategies than others, we assign higher importance to pairs whose choice of strategy matters more.  
We therefore adopt a \emph{weighted} variant of Spearman’s correlation in which each item is given a non-negative weight \(w_i\).

\paragraph{Definition.}
Let \(R_{1,i}\) and \(R_{2,i}\) be the ranks of the \(i\)-th item under two orderings and let \(w_i\) be its weight.  
Denote the weighted means
\[
\bar{R}_1=\frac{\sum_{i=1}^n w_i R_{1,i}}{\sum_{i=1}^n w_i},
\qquad
\bar{R}_2=\frac{\sum_{i=1}^n w_i R_{2,i}}{\sum_{i=1}^n w_i}.
\]
The \textbf{weighted Spearman correlation} is then the weighted Pearson correlation between the rank vectors:
{\small
\[
\rho_w \;=\;
\frac{\displaystyle\sum_{i=1}^n w_i\,(R_{1,i}-\bar{R}_1)\,(R_{2,i}-\bar{R}_2)}
     {\sqrt{\displaystyle\sum_{i=1}^n w_i\,(R_{1,i}-\bar{R}_1)^2}\;
      \sqrt{\displaystyle\sum_{i=1}^n w_i\,(R_{2,i}-\bar{R}_2)^2}}.
\]
}
\paragraph{Choice of weights.}
For each \{model,\,dataset\} pair, we first train the target model on data produced by every candidate response-generation method and record the resulting accuracies.  
The weight \(w_i\) is set to the \emph{standard deviation} of these accuracies.
Intuitively, tasks in which the strategies yield very different outcomes (\(w_i\) large) are more informative when judging a ranking metric, so they contribute more to $\rho_w$.

\paragraph{Interpretation.}
A value of $\rho_w\!\approx\!1$ indicates that the metric produces a ranking almost identical to the gold ranking, with higher-variance tasks influencing the score most strongly.  
Conversely, $\rho_w\!\approx\!0$ implies no weighted monotonic relationship, and $\rho_w\!\approx\!-1$ signals an inverse agreement.

Throughout the main text and Appendix, all reported “Spearman” results actually correspond to this weighted formulation.

\subsection{\textbf{Data Selection Rationale for the Benchmark}}

The datasets included in our benchmark, drawn from the Mistral, Llama, and Qwen benchmarks, were selected according to a specific set of rules designed to ensure relevance and suitability.  These rules are as follows:

1.  Sufficient Dataset Size: We only included datasets where the combined size of the training, validation, and testing sets exceeded 650 samples. This threshold was chosen to ensure sufficient data for robust model evaluation.

2.  Accuracy as Evaluation Metric: A key requirement was that the dataset could be evaluated using accuracy as the primary metric. This allows for a clear and quantifiable assessment of model performance.

3.  English Question-Answering Format:  All selected datasets are in an English question-and-answer format to maintain consistency and focus on English language reasoning abilities.

4.  Focus on Reasoning Tasks: The underlying task presented by each dataset must involve reasoning skills.  This ensures that the benchmark effectively assesses the models' ability to reason and infer.

A detailed justification for the inclusion or exclusion of each dataset can be found in Table~\ref{tab:why_each_data_is_chosen}.

\subsection{Correctness Filter}\label{sec:select_correct_examples}

Without supervised fine-tuning (SFT), $M(x)$ may generate incorrect responses, making cosine similarity calculations between $M(x)$and $\hat{y}$ unreliable.
To alleviate this, we introduce a filtering mechanism to filter out the incorrect $M(x)$. We notice that for mathematical problems, the correct final answer typically appears as the last number in $M(x)$. Therefore, for Math-related tasks, we use regular expressions (regex) to extract the last number from the prediction and compare it directly with the ground truth. For other types of problem, we use the Qwen2.5-Instruct 7b model to extract the predicted label from the model output. We then compare this extracted label with the true gold label; if they match, we consider the prediction correct by default.

\begin{table*}[t]
  \centering
  \small
  \begin{tabular}{l|l|c|c|c|c|c|c}
    \hline
    Methods & STD Range & \#Data & Mistral & Llama & Qwen & Acc & Weighted $\rho$ \\
    \hline
    Ours - Ours w/o Filter & All        & 51 & $-0.01\%$ & $-0.03\%$ & $0.00\%$  & $-0.02\%$ & $+0.007$ \\
    Ours - Ours w/o Filter & $\text{STD}>2\%$ & 27 & $+0.01\%$ & $-0.03\%$ & $+0.37\%$ & $+0.03\%$ & $+0.013$ \\
    Ours - Ours w/o Filter & $\text{STD}>4\%$ & 14 & $+0.01\%$ & $-0.10\%$ & $+0.55\%$ & $+0.05\%$ & $+0.012$ \\
    \hline
  \end{tabular}
  \caption{Comparison between our method and our method without the correctness filter across different variance ranges.}
  \label{tab:ours_vs_wofilter}
\end{table*}

\begin{table*}[t]
  \centering
  \small
  \begin{tabular}{l|l|c|c|c|c|c|c}
    \hline
    Methods & STD Range & \#Data & Mistral & Llama & Qwen & Acc & Weighted $\rho$ \\
    \hline
    Ours - Ours w/o Filter & All Data & 18.0 & $+2.97\%$ & $+3.21\%$ & $+5.52\%$ & $+3.90\%$ & $+0.068$ \\
    \hline
  \end{tabular}
  \caption{Performance gain of our method compared with the variant without the correctness filter.}
  \label{tab:ours_vs_wofilter_single}
\end{table*}

\subsubsection*{Response Construction Details}
\label{appendix:Response Construction Details}

\textbf{Ground Truth}: This strategy uses the original ground-truth responses from the datasets as target outputs. Since our focus is on selecting effective chain-of-thought (CoT) target responses, we apply this method to datasets that include human-annotated CoT reasoning steps, such as GSM8K, MATH, ECQA, MBPP. When human-annotated CoT is unavailable, we use the gold label as ground truth.

\textbf{\hyperref[prompt answer directly]{GPT-4o Answer Directly}}, \textbf{\hyperref[prompt answer directly]{Claud Answer Directly}}, and \textbf{\hyperref[prompt answer directly]{MiniGPT-4o Answer Directly}} generate responses based on questions and the ground truth using GPT-4o, Claude 3.5 and Mini-GPT4, respectively. \textbf{\hyperref[prompt rewrite groundtruth]{Rewrite Ground Truth:}} Direct GPT-4o to restyle the ground truth in its own language. This method is only applicable to GSM8K, MATH Algebra, ECQA. The other tasks's ground truth consists of target labels without any human-annotated chain-of-thought (CoT) reasoning, making rewriting infeasible. \textbf{\hyperref[prompt step by step]{Step-by-Step:}} instructs GPT-4o to generate step-by-step responses based on questions and ground truth. \textbf{\hyperref[prompt gpt4o example]{GPT-4o Examples:}} To facilitate problem-solving, we provide GPT-4o with two high-quality, expert-selected in-context examples of its own responses. GPT-4o is then tasked with generating new responses based on these examples. \textbf{\hyperref[prompt human written example]{Human Examples:}} To aid GPT-4o in understanding problem-solving for these datasets, we provide two carefully chosen human-written examples as context. GPT-4o then uses these examples to generate new responses. We put more details in Section~\ref{sect:gpt4_varient_appendix} in Appendix.

\subsubsection{Prompt for Self-Aligned Perplexity}

\subsubsection*{Redundant Prompt}
\label{appendix:redundant_prompt}
\begin{figure*}[ht]
\centering
\includegraphics[width=0.95\textwidth]{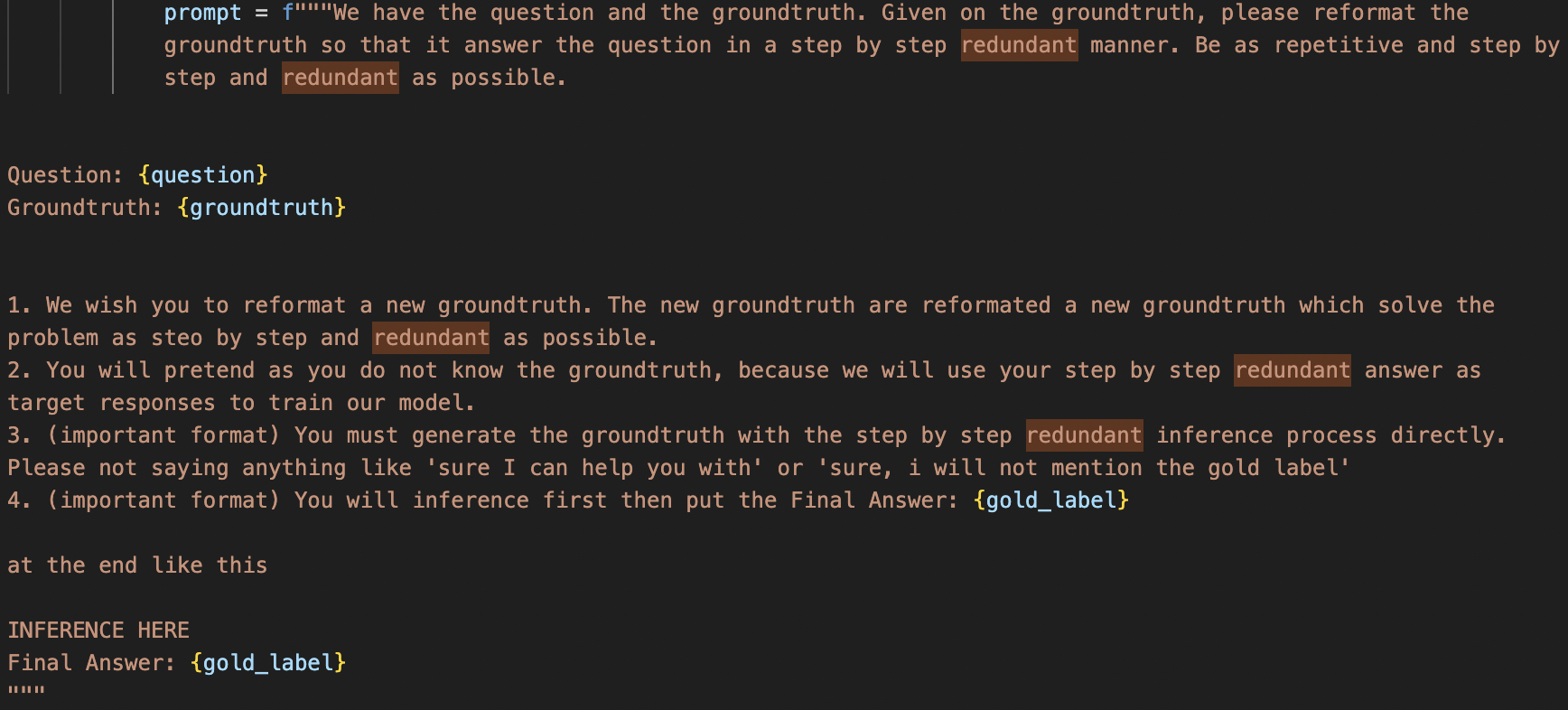}
\caption{Prompt that we used for generate step by step responses}
\label{figure:redundant}
\end{figure*}

We construct redundant prompts (shown in the Figure~\ref{figure:redundant}) to demonstrate that the perplexity of the redundant target responses is lower than that of GPT-4's answers. Perplexity primarily reflects how fluent the language is and how well the language style aligns with the model, but it places less emphasis on semantic meaning.

\begin{figure*}[ht]
\centering
\includegraphics[width=0.95\textwidth]{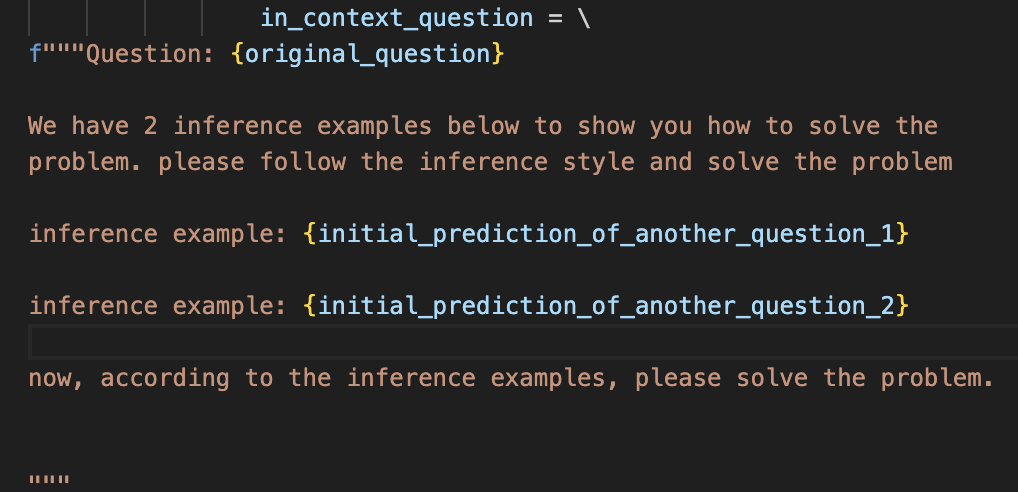}
\caption{Prompt that we used for self-aligned perplexity}
\label{appendix:self_aligned_ppl_prompt}
\end{figure*}

\textbf{Self-Aligned In-Context Prompt for Perplexity Calculation}\label{in_context_prompt}
The prompt shown from the Figure~\ref{appendix:self_aligned_ppl_prompt} shows how we add self - generated initial predictions from other questions as in - context examples for perplexity calculation.

\subsubsection{Data Generation Strategies}\label{sect:gpt4_varient_appendix}
We instruct GPT-4, Claude 3.5, Mini-GPT4 to generate different of target responses using different target reponse generation strategies.

\textbf{GPT-4/Claude 3.5/Mini-GPT4 Answer Directly:}\label{prompt answer directly} This prompt is from \citep{ren2024learn}. For tasks involving mathematics and coding, we submit the problems from our training dataset directly to GPT-4 or Claude 3.5 to obtain their solutions. In the case of classification tasks, we provide these models with the input questions alongside the correct labels (excluding any human-generated explanations) and utilize their outputs. These generated answers are then paired with the original questions to form the GPT-4/Claude 3.5 Direct Answer Training Dataset.

To ensure that the models develop their own problem-solving and analytical capabilities, we deliberately exclude any solutions or rationales related to math, coding, or classification tasks. This approach prevents the models from simply mimicking the ground truth processes, which could otherwise result in some of GPT-4’s predictions lacking its unique reasoning style. Such mimicry would undermine the reliability of our perplexity measurements, which are designed to evaluate how effectively a language model handles outputs from other models.

\begin{figure*}[ht]
\centering
\includegraphics[width=0.95\textwidth]{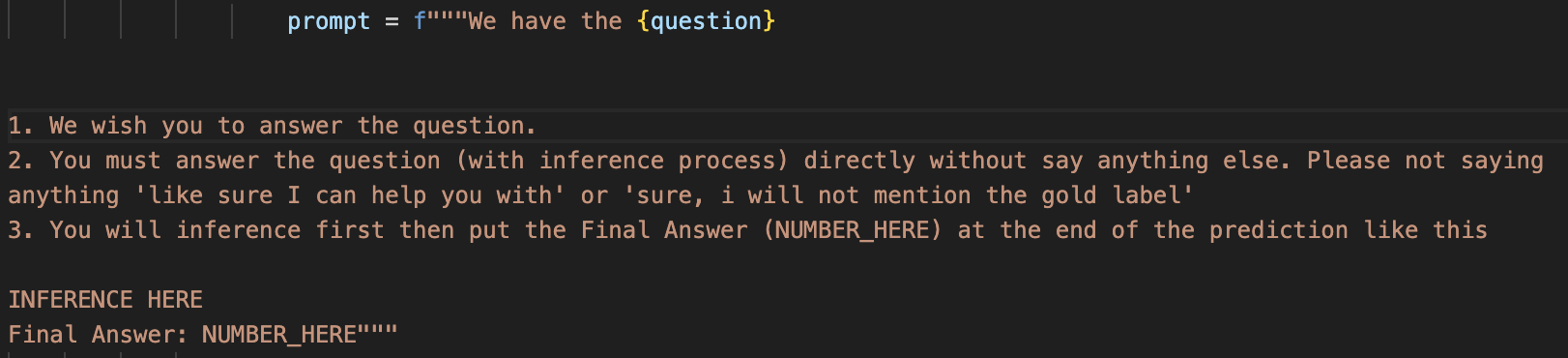}
\caption{Prompt that we used for generate GPT-4/Claude 3.5/Mini-GPT4 Answer Directly responses}
\label{figure:answer}
\end{figure*}

The prompt from the Figure~\ref{figure:answer} below is designed to guide GPT-4/Claude 3.5 in generating responses without relying on the ground truth solutions:

\textbf{Rewrite Ground Truth:}\label{prompt rewrite groundtruth} This prompt is from \citep{ren2024learn}. In this approach, we provide GPT-4 and Claude 3.5 with the ground truth data, which includes human-annotated rationales and detailed problem-solving steps. The goal is to have GPT-4 and Claude 3.5 rephrase the ground truth content using their own linguistic styles.

\begin{figure*}[ht]
\centering
\includegraphics[width=0.95\textwidth]{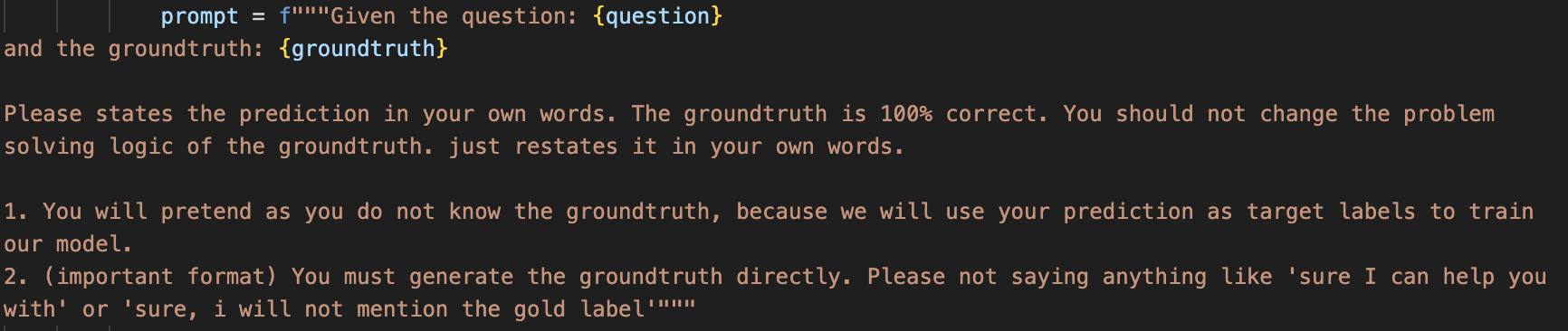}
\caption{Prompt that we used for generate the Rewrite Ground Truth style responses}
\label{figure:rewrite}
\end{figure*}

The subsequent prompt(shown in the Figure~\ref{figure:rewrite}) guides GPT-4 and Claude 3.5 to generate the GPT-4/Claude 3.5 Response (Rewrite GT) output.

\textbf{Step-by-step:}\label{prompt step by step} This prompt is from \citep{ren2024learn}. We instruct GPT-4 and Claude 3.5 to methodically address each problem by breaking it down into sequential steps. For tasks involving mathematics and coding, we present the problems directly from our training dataset to these models to obtain their solutions. In classification tasks, we provide GPT-4 and Claude 3.5 with the correct labels (excluding any human-generated explanations) along with the input questions, and then utilize their detailed, step-by-step responses. These generated answers are subsequently paired with the original questions to form the GPT-4/Claude 3.5 Step-by-Step Response (No GT) Dataset.

To ensure that the models develop their own unique problem-solving and analytical approaches, we intentionally exclude the solutions or rationales for the mathematics, coding, or classification tasks. This prevents the models from simply mimicking the problem-solving and analytical methods found in the ground truth data. Including such processes could result in some of GPT-4’s and Claude 3.5’s outputs not reflecting their inherent reasoning styles, thereby compromising the accuracy of our perplexity measurements. These measurements are designed to assess how effectively a language model can handle outputs generated by other language models.

\begin{figure*}[ht]
\centering
\includegraphics[width=0.95\textwidth]{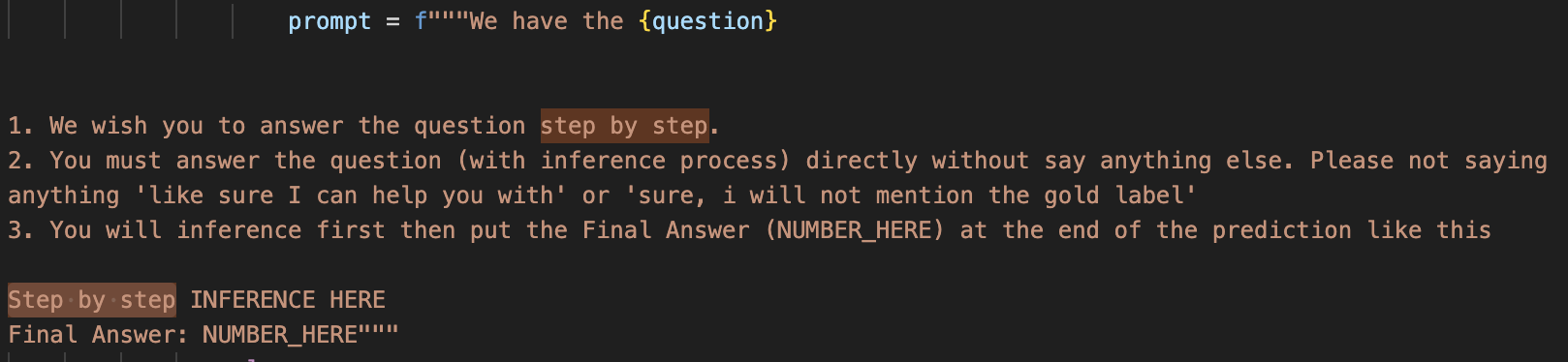}
\caption{Prompt that we used for generate step by step responses}
\label{figure:step}
\end{figure*}

The following prompt from the Figure~\ref{figure:step} directs GPT-4 and Claude 3.5 to generate the GPT-4/Claude 3.5 Step-by-Step Response (No GT) responses.

\textbf{GPT-4o with GPT-4o Examples:} \label{prompt gpt4o example} We developed this prompt specifically for the API-Bank and Plan-Bench datasets. This prompt utilizes GPT-4's own accurate generations as examples to help GPT-4 not only better understand the task but also demonstrate how to solve the problems effectively. The prompt below is an example that we used to generate target responses for the API-Bank dataset.

\begin{figure*}[ht]
\centering
\includegraphics[width=0.95\textwidth]{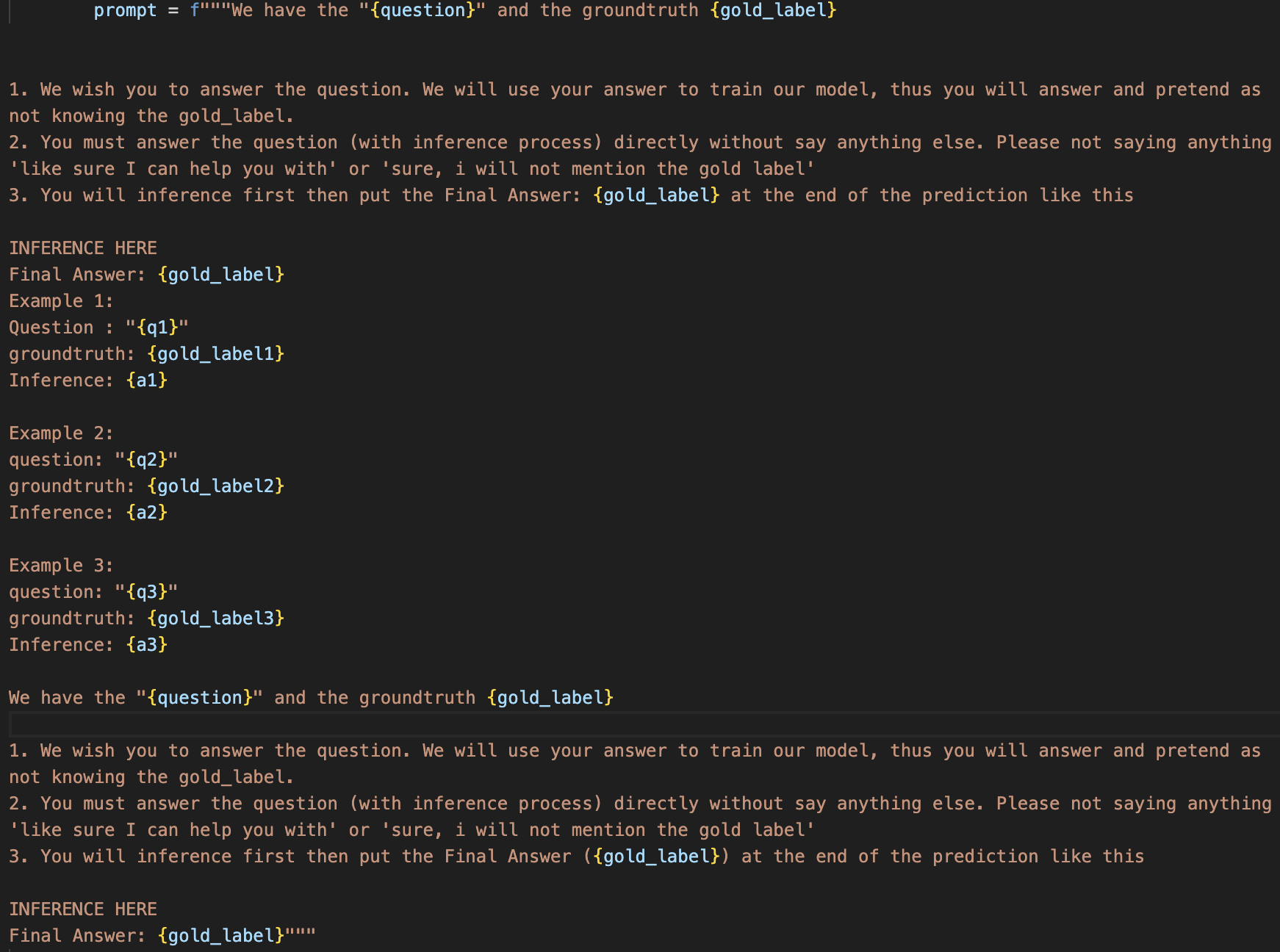}
\caption{Prompt that we used for generate responses guided by GPT-4o with GPT-4o/Human written Examples}
\label{figure:example}
\end{figure*}

The following prompt from the Figure~\ref{figure:example} directs GPT-4o to generate responses guided by GPT-4o generated Examples

\textbf{GPT-4 with Human Written Examples:}\label{prompt human written example} We developed this prompt specifically for the API-Bank and Plan-Bench datasets. This prompt utilizes human written examples to help GPT-4 not only better understand the task but also demonstrate how to solve the problems effectively. The prompt below is an example that we used to generate target responses for the API-Bank dataset.

The following prompt from the Figure~\ref{figure:example} directs GPT-4o to generate responses guided by Human written Examples

\subsection{AI Assistant}
We used GPT-4o as a writing assistant and programming aid for editing purposes.

\subsection{Required Compute Resources}
Each individual training run reported in this paper requires approximately 5–48 GPU hours when using a 40GB A100 GPU. We do not recommend you to reproduce every training run, as there are too many experiments. Instead, we strongly recommend directly using the reported training outcomes from each table as the final results. You can then compute your ranking metrics to evaluate how well your metric aligns with the training outcomes. Calculating metrics such as perplexity on a small subset of all of the dataset takes only about 2 hours on a single 40GB A100 GPU.

\begin{table*}[ht]
\centering
\resizebox{\textwidth}{!}{
\begin{tabular}{l|l|l|c|c|c|c|c|c|c|c|}
\hline

Method & Model Type & training task& GSM8K & Math Algebra  & ECQA &SQUAD & DROP & WINOGRANDE \\ \hline

Gold Label & Mistral& ECQA& 0.383 &0.181  &  \cellcolor{gray!50}0.722 & 0.251 & 0.084&  0.562 \\
GPT-4o Answer Directly &  &  & 0.484          & 0.218 & \cellcolor{gray!50} 0.707 & 0.175& 0.016 & 0.638  \\
\hline
Gold Label & Mistral& SQUAD& 0.082 & 0.0931 &  0.633 & \cellcolor{gray!50} 0.74 &0.208 & 0.566 \\
GPT-4o Answer Directly& & &0.512  & 0.234 &  0.594 & \cellcolor{gray!50}0.748 & 0.268&  0.628 \\
\hline
Gold Label & Mistral& DROP& 0.076 &0.097 & 0.621  &  0.561&\cellcolor{gray!50} 0.628 & 0.578 \\
GPT-4o Answer Directly &      &       & 0.542 & 0.241 &0.602  & 0.546 & \cellcolor{gray!50} 0.736 & 0.638 \\
\hline
Gold Label & Mistral& WINOGRANDE& 0.381 & 0.172 & 0.625  & 0.166 & 0.042 &\cellcolor{gray!50} 0.742 \\
GPT-4o Answer Directly &      &        & 0.477 & 0.219 & 0.569  & 0.106 & 0.016& \cellcolor{gray!50}0.713 \\

\hline
\hline
\hline

Gold Label & LLAMA3& ECQA& 0.798 & 0.416&  \cellcolor{gray!50}0.734 & 0.193 &0.1 &  0.637\\
GPT-4o Answer Directly & &  &  0.778           &0.469  & \cellcolor{gray!50} 0.723 &0.389  & 0.284 &0.638 \\
\hline
Gold Label & LLAMA3& SQUAD& 0.584 & 0.366 &  0.712 & \cellcolor{gray!50} 0.758& 0.49&0.639  \\
GPT-4o Answer Directly &      &         & 0.791 & 0.457 &  0.726 & \cellcolor{gray!50}0.759 & 0.368& 0.651 \\
\hline
Gold Label & LLAMA3& DROP& 0.144 & 0.169 &  0.674 &  0.574&\cellcolor{gray!50} 0.738 & 0.582  \\
GPT-4o Answer Directly  &     &        &  0.776& 0.507 & 0.703 & 0.555 & \cellcolor{gray!50} 0.786 & 0.626 \\
\hline
Gold Label & LLAMA3& WINOGRANDE&  0.776& 0.445 &  0.717 &  0.226& 0.162&\cellcolor{gray!50} 0.766 \\
GPT-4o Answer Directly &        &      & 0.775 & 0.485 &  0.721 & 0.305 & 0.238 & \cellcolor{gray!50} 0.695 \\

\hline
\hline
\hline

Gold Label & Qwen& ECQA&0.914  & 0.903 &  \cellcolor{gray!50}0.814 &   0.662& 0.008 & 0.675  \\
GPT-4o Answer Directly& &  &    0.903         &  0.888& \cellcolor{gray!50} 0.793 & 0.668 & 0.016& 0.716 \\
\hline
Gold Label & Qwen& SQUAD & 0.899 &  0.892 &  0.784& \cellcolor{gray!50} 0.768 & 0.056 & 0.693  \\
GPT-4o Answer Directly &     &          &0.896  &  0.911&  0.789 & \cellcolor{gray!50} 0.756 & 0.074&0.712   \\
\hline
Gold Label & Qwen& DROP& 0.788 &0.904 & 0.799  &  0.701&\cellcolor{gray!50}0.664  & 0.711  \\
GPT-4o Answer Directly &     &        & 0.911 & 0.903 & 0.792 & 0.741 & \cellcolor{gray!50} 0.806 & 0.701 \\
\hline
Gold Label & Qwen& WINOGRANDE & 0.893 & 0.904  & 0.78 & 0.651 & 0.004&\cellcolor{gray!50} 0.725 \\
GPT-4o Answer Directly&       &       & 0.902 & 0.896 & 0.798  &  0.68&0.022 & \cellcolor{gray!50} 0.721 \\
\hline

\hline

\end{tabular}

}
\caption{The training data size is 1000. This table compares the in-domain and cross-domain performance when training on gold-label vs. GPT-4 generated synthetic data. As can be seen from the table, the in-domain performance of the model is typically higher when training with gold-label data. However, the cross-domain performance when training on GPT-4 generated data is significantly higher than when training with only gold-label data. The grey area represents the in-domain performance.}

\label{table:cross_domain_performance_recording}

\end{table*}

\begin{table*}[t!]
\resizebox{1.0\textwidth}{!}{
\begin{tabular}{l|l|c|c|c}
\hline
Dataset&\textbf{Method} & \textbf{Accuracy and N train} & \textbf{Mistral} &\textbf{Llama3-8B-Chat} \\ \hline 
\textbf{MATH Algebra} &GPT4 preview&  82.5\%, 1000&   0.301& 0.504\\ 
 &GPT4 only correct & 100\%, 825&0.293 &  0.501\\ 
 &GPT4 only correct + rewritten ground truth &    100\%, 1000 & 0.293 & 0.500\\ 
\hline
\textbf{MATH Algebra} &Claude& 90.1\%, 1000&  0.265&  0.508\\ 
 &Claude only correct &  100\%, 901 &  0.277&  0.487\\ 
 &Claude only correct + rewritten ground truth &     100\%, 1000& 0.286& 0.492\\ 
\hline
\textbf{MATH Algebra} &Mini GPT4& 91.6\%  , 1000   & 0.313 &0.523\\ 
 &Mini GPT4 only correct &   100\%, 916 & 0.311 & o.523\\ 
 &Mini GPT4 only correct + rewritten ground truth &   100\%, 1000 & 0.326& 0.539\\ 
\hline

\hline
 
\textbf{GSM8K} &GPT4 preview&  92.1\%, 1000&   0.597& 0.799\\ 
 &GPT4 only correct & 100\%, 921& 0.587& 0.791 \\ 
 &GPT4 only correct + rewritten ground truth &    100\%, 1000 & 0.588& 0.808\\ 
\hline
\textbf{GSM8K} &Claude& 95.6\%, 1000& 0.578 & 0.796 \\ 
&Claude only correct &  100\%, 956 & 0.580 & 0.797  \\ 
 &Claude only correct + rewritten ground truth &     100\%, 1000& 0.588&0.798 \\ 
\hline
\textbf{GSM8K} &Mini GPT4& 89.8\%  , 1000   & 0.623 & 0.795\\ 
 &Mini GPT4 only correct &   100\%, 898  & 0.606 & 0.793\\ 
 &Mini GPT4 only correct + rewritten ground truth &   100\%, 1000 &0.607 &0.790 \\ 
\hline
\end{tabular}
}

\caption{The table shows that the accuracy of the generated data has a marginal effect on the training outcome. In this table, we use the API with different math abilities. The rank of their math problem-solving abilities is: Claude \textgreater MiniGPT-4 \textgreater GPT-4 preview. GPT-4 preview represents the data generated using the GPT-4 preview model, rather than the GPT-4o model.}
\label{tab:impact_accuracy}
\end{table*}

\begin{table*}[t!]
\centering
\resizebox{1.0\textwidth}{!}{
\begin{tabular}{l|l|c|c|c|c|c|c|c|c|c|c|c|c|c|c|c|c}
\hline
Data Generation Strategy & Model Type & gsm8k & math algebra & math geometry & ecqa & boolq & winogrande & piqa & agieval & squad & arc challenge & drop & mbpp & api bank & hellaswag & mmlu pro law & mmlu moral scenarios \\ \hline
gold label & mistral &  &  &  & 0.717 & 0.997 & 0.736 & 0.854 & 0.44 & 0.741 & 0.747 & 0.645 &  & 0.452 & 0.772 & 0.263 & 0.679\\
groundtruth &  & 0.442 & 0.194 & 0.125 & 0.684 &  &  &  &  &  &  &  & 0.325 &  &  &  & \\
gpt4 &  & 0.62 & 0.324 & 0.146 & 0.703 & 0.87 & 0.717 & 0.864 & 0.41 & 0.732 & 0.631 & 0.723 & 0.362 & 0.515 & 0.659 & 0.238 & 0.691\\
claude &  & 0.582 & 0.278 & 0.136 & 0.735 & 0.885 & 0.724 & 0.848 & 0.445 & 0.736 & 0.753 & 0.729 & 0.379 & 0.579 & 0.553 & 0.248 & 0.751\\
mini gpt4 &  & 0.619 & 0.306 & 0.151 & 0.708 & 0.882 & 0.695 & 0.868 & 0.427 & 0.732 & 0.772 & 0.735 & 0.348 & 0.43 & 0.663 & 0.205 & 0.659\\
step by step &  & 0.626 & 0.314 & 0.137 & 0.706 & 0.874 & 0.693 & 0.862 & 0.445 & 0.749 & 0.71 & 0.696 & 0.333 & 0.377 & 0.644 & 0.249 & 0.714\\
openai human written examples &  & 0.621 & 0.303 & 0.163 & 0.708 & 0.891 & 0.721 & 0.859 & 0.413 & 0.76 & 0.692 & 0.741 & 0.345 & 0.411 & 0.674 & 0.233 & 0.71\\
gpt4 style in context examples &  & 0.61 & 0.254 & 0.158 & 0.726 & 0.884 & 0.727 & 0.868 & 0.44 & 0.761 & 0.697 & 0.735 & 0.378 & 0.416 & 0.672 & 0.225 & 0.728\\
rewrite groundtruth in own words &  & 0.502 & 0.238 & 0.127 & 0.703 &  &  &  &  &  &  &  & 0.306 &  &  &  & \\
\hline\hline
gold label & llama 3 instruct &  &  &  & 0.737 & 0.979 & 0.761 & 0.852 & 0.432 & 0.756 & 0.766 & 0.742 &  & 0.507 & 0.772 & 0.332 & 0.639\\
groundtruth &  & 0.678 & 0.404 & 0.239 & 0.701 &  &  &  &  &  &  &  & 0.445 &  &  &  & \\
gpt4 &  & 0.816 & 0.559 & 0.301 & 0.74 & 0.87 & 0.697 & 0.866 & 0.448 & 0.759 & 0.806 & 0.793 & 0.482 & 0.477 & 0.712 & 0.247 & 0.659\\
claude &  & 0.803 & 0.5 & 0.254 & 0.756 & 0.865 & 0.72 & 0.86 & 0.445 & 0.765 & 0.801 & 0.757 & 0.471 & 0.547 & 0.709 & 0.259 & 0.737\\
mini gpt4 &  & 0.805 & 0.551 & 0.28 & 0.721 & 0.864 & 0.677 & 0.868 & 0.437 & 0.747 & 0.816 & 0.783 & 0.491 & 0.384 & 0.719 & 0.225 & 0.645\\
step by step &  & 0.797 & 0.562 & 0.26 & 0.731 & 0.869 & 0.72 & 0.853 & 0.433 & 0.779 & 0.792 & 0.78 & 0.455 & 0.227 & 0.71 & 0.242 & 0.684\\
openai human written examples &  & 0.81 & 0.547 & 0.283 & 0.735 & 0.893 & 0.717 & 0.867 & 0.44 & 0.766 & 0.804 & 0.807 & 0.477 & 0.347 & 0.706 & 0.229 & 0.667\\
gpt4 style in context examples &  & 0.796 & 0.494 & 0.285 & 0.736 & 0.885 & 0.719 & 0.87 & 0.447 & 0.752 & 0.811 & 0.794 & 0.47 & 0.368 & 0.721 & 0.259 & 0.681\\
rewrite groundtruth in own words &  & 0.742 & 0.444 & 0.241 & 0.727 &  &  &  &  &  &  &  & 0.437 &  &  &  & \\
\hline\hline
gold label & qwen &  &  &  & 0.816 & 0.892 & 0.732 & 0.867 & 0.48 & 0.77 & 0.855 & 0.663 &  & 0.515 & 0.74 & 0.303 & 0.605\\
groundtruth &  & 0.899 & 0.894 & 0.667 & 0.793 &  &  &  &  &  &  &  & 0.59 &  &  &  & \\
gpt4 &  & 0.897 & 0.916 & 0.679 & 0.794 & 0.858 & 0.709 & 0.878 & 0.552 & 0.76 & 0.886 & 0.798 & 0.591 & 0.436 & 0.722 & 0.3 & 0.656\\
claude &  & 0.895 & 0.904 & 0.648 & 0.788 & 0.862 & 0.72 & 0.88 & 0.553 & 0.766 & 0.874 & 0.793 & 0.607 & 0.462 & 0.72 & 0.309 & 0.66\\
mini gpt4 &  & 0.904 & 0.904 & 0.654 & 0.787 & 0.87 & 0.712 & 0.882 & 0.555 & 0.763 & 0.891 & 0.821 & 0.642 & 0.379 & 0.701 & 0.308 & 0.664\\
step by step &  & 0.899 & 0.907 & 0.642 & 0.792 & 0.859 & 0.716 & 0.88 & 0.548 & 0.767 & 0.881 & 0.806 & 0.623 & 0.417 & 0.713 & 0.287 & 0.662\\
openai human written examples &  & 0.905 & 0.909 & 0.647 & 0.787 & 0.871 & 0.697 & 0.883 & 0.547 & 0.794 & 0.883 & 0.82 & 0.628 & 0.458 & 0.724 & 0.283 & 0.608\\
gpt4 style in context examples &  & 0.899 & 0.903 & 0.657 & 0.803 & 0.88 & 0.731 & 0.878 & 0.57 & 0.785 & 0.868 & 0.809 & 0.631 & 0.309 & 0.742 & 0.3 & 0.642\\
rewrite groundtruth in own words &  & 0.902 & 0.904 & 0.654 & 0.787 &  &  &  &  &  &  &  & 0.589 &  &  &  & \\
\hline
\end{tabular}}
\caption{average of seed 0,1,2 train datasize 1000 lr 2e-05 epoch num 20}
\label{tab:ntrain_1000_lr_2e-05_seedaverage_of_seed_0,1,2}
\end{table*}

\begin{table*}[t!]
\centering
\resizebox{1.0\textwidth}{!}{
\begin{tabular}{l|l|c|c|c|c|c|c|c|c|c|c|c|c|c|c|c|c}
\hline
Data Generation Strategy & Model Type & gsm8k & math algebra & math geometry & ecqa & boolq & winogrande & piqa & agieval & squad & arc challenge & drop & mbpp & api bank & hellaswag & mmlu pro law & mmlu moral scenarios \\ \hline
gold label & mistral &  &  &  & 0.722 & 0.996 & 0.742 & 0.852 & 0.440 & 0.748 & 0.759 & 0.628 &  & 0.465 & 0.771 & 0.252 & 0.650\\
groundtruth &  & 0.440 & 0.201 & 0.110 & 0.672 &  &  &  &  &  &  &  & 0.370 &  &  &  & \\
gpt4 &  & 0.625 & 0.319 & 0.177 & 0.700 & 0.867 & 0.713 & 0.869 & 0.400 & 0.732 & 0.611 & 0.746 & 0.347 & 0.510 & 0.654 & 0.229 & 0.713\\
claude &  & 0.583 & 0.279 & 0.160 & 0.720 & 0.886 & 0.709 & 0.849 & 0.425 & 0.728 & 0.732 & 0.726 & 0.403 & 0.584 & 0.549 & 0.219 & 0.760\\
mini gpt4 &  & 0.627 & 0.291 & 0.148 & 0.710 & 0.873 & 0.688 & 0.877 & 0.420 & 0.740 & 0.775 & 0.726 & 0.363 & 0.433 & 0.663 & 0.183 & 0.643\\
step by step &  & 0.639 & 0.323 & 0.127 & 0.705 & 0.885 & 0.687 & 0.861 & 0.445 & 0.752 & 0.708 & 0.676 & 0.340 & 0.478 & 0.639 & 0.196 & 0.723\\
openai human written examples &  & 0.604 & 0.306 & 0.160 & 0.709 & 0.897 & 0.718 & 0.869 & 0.420 & 0.756 & 0.685 & 0.742 & 0.350 & 0.400 & 0.664 & 0.196 & 0.717\\
gpt4 style in context examples &  & 0.619 & 0.231 & 0.169 & 0.725 & 0.887 & 0.732 & 0.879 & 0.430 & 0.764 & 0.678 & 0.732 & 0.373 & 0.433 & 0.687 & 0.223 & 0.710\\
rewrite groundtruth in own words &  & 0.511 & 0.231 & 0.127 & 0.709 &  &  &  &  &  &  &  & 0.323 &  &  &  & \\
\hline\hline
gold label & llama 3 instruct &  &  &  & 0.734 & 0.978 & 0.766 & 0.855 & 0.435 & 0.761 & 0.764 & 0.738 &  & 0.502 & 0.777 & 0.312 & 0.630\\
groundtruth &  & 0.681 & 0.396 & 0.215 & 0.691 &  &  &  &  &  &  &  & 0.450 &  &  &  & \\
gpt4 &  & 0.814 & 0.562 & 0.278 & 0.723 & 0.880 & 0.695 & 0.865 & 0.435 & 0.752 & 0.801 & 0.796 & 0.480 & 0.494 & 0.722 & 0.276 & 0.677\\
claude &  & 0.816 & 0.493 & 0.253 & 0.748 & 0.879 & 0.728 & 0.864 & 0.455 & 0.763 & 0.808 & 0.746 & 0.500 & 0.547 & 0.710 & 0.286 & 0.757\\
mini gpt4 &  & 0.795 & 0.557 & 0.278 & 0.725 & 0.867 & 0.702 & 0.863 & 0.450 & 0.739 & 0.826 & 0.730 & 0.500 & 0.384 & 0.703 & 0.223 & 0.670\\
step by step &  & 0.798 & 0.564 & 0.308 & 0.728 & 0.874 & 0.718 & 0.866 & 0.460 & 0.783 & 0.792 & 0.780 & 0.450 & 0.216 & 0.715 & 0.229 & 0.657\\
openai human written examples &  & 0.811 & 0.547 & 0.266 & 0.736 & 0.891 & 0.719 & 0.864 & 0.450 & 0.770 & 0.809 & 0.808 & 0.457 & 0.355 & 0.699 & 0.269 & 0.640\\
gpt4 style in context examples &  & 0.792 & 0.515 & 0.274 & 0.742 & 0.875 & 0.717 & 0.854 & 0.460 & 0.755 & 0.809 & 0.798 & 0.483 & 0.273 & 0.718 & 0.219 & 0.683\\
rewrite groundtruth in own words &  & 0.729 & 0.443 & 0.241 & 0.715 &  &  &  &  &  &  &  & 0.417 &  &  &  & \\
\hline\hline
gold label & qwen &  &  &  & 0.814 & 0.880 & 0.725 & 0.868 & 0.500 & 0.769 & 0.856 & 0.652 &  & 0.518 & 0.747 & 0.296 & 0.590\\
groundtruth &  & 0.906 & 0.898 & 0.675 & 0.784 &  &  &  &  &  &  &  & 0.610 &  &  &  & \\
gpt4 &  & 0.889 & 0.916 & 0.658 & 0.793 & 0.865 & 0.721 & 0.879 & 0.545 & 0.762 & 0.890 & 0.794 & 0.607 & 0.433 & 0.706 & 0.276 & 0.633\\
claude &  & 0.884 & 0.906 & 0.662 & 0.796 & 0.873 & 0.716 & 0.885 & 0.550 & 0.767 & 0.867 & 0.798 & 0.600 & 0.457 & 0.717 & 0.322 & 0.667\\
mini gpt4 &  & 0.905 & 0.904 & 0.654 & 0.782 & 0.865 & 0.704 & 0.881 & 0.535 & 0.760 & 0.891 & 0.818 & 0.633 & 0.396 & 0.704 & 0.299 & 0.657\\
step by step &  & 0.899 & 0.908 & 0.624 & 0.795 & 0.846 & 0.703 & 0.874 & 0.545 & 0.752 & 0.882 & 0.766 & 0.630 & 0.412 & 0.717 & 0.276 & 0.653\\
openai human written examples &  & 0.907 & 0.910 & 0.658 & 0.790 & 0.876 & 0.699 & 0.884 & 0.540 & 0.808 & 0.881 & 0.816 & 0.617 & 0.445 & 0.731 & 0.286 & 0.583\\
gpt4 style in context examples &  & 0.896 & 0.902 & 0.654 & 0.799 & 0.883 & 0.734 & 0.871 & 0.540 & 0.782 & 0.863 & 0.800 & 0.607 & 0.339 & 0.742 & 0.302 & 0.653\\
rewrite groundtruth in own words &  & 0.911 & 0.899 & 0.654 & 0.791 &  &  &  &  &  &  &  & 0.587 &  &  &  & \\
\hline
\end{tabular}}
\caption{seed 0 train datasize 1000 lr 2e-05 epoch num 20}
\label{tab:ntrain_1000_lr_2e-05_seed0}
\end{table*}

\begin{table*}[t!]
\centering
\resizebox{1.0\textwidth}{!}{
\begin{tabular}{l|l|c|c|c|c|c|c|c|c|c|c|c|c|c|c|c|c}
\hline
Data Generation Strategy & Model Type & gsm8k & math algebra & math geometry & ecqa & boolq & winogrande & piqa & agieval & squad & arc challenge & drop & mbpp & api bank & hellaswag & mmlu pro law & mmlu moral scenarios \\ \hline
gold label & mistral &  &  &  & 0.714 & 0.997 & 0.733 & 0.855 &  & 0.738 & 0.741 & 0.654 &  & 0.445 & 0.772 & 0.269 & 0.693\\
groundtruth &  & 0.443 & 0.191 & 0.131 & 0.690 &  &  &  &  &  &  &  & 0.303 &  &  &  & \\
gpt4 &  & 0.617 & 0.327 & 0.148 & 0.704 & 0.872 & 0.719 & 0.861 & 0.415 & 0.732 & 0.641 & 0.712 & 0.370 & 0.518 & 0.662 & 0.243 & 0.680\\
claude &  & 0.581 & 0.277 & 0.143 & 0.742 & 0.885 & 0.731 & 0.847 & 0.455 & 0.740 & 0.764 & 0.730 & 0.367 & 0.576 & 0.555 & 0.262 & 0.747\\
mini gpt4 &  & 0.615 & 0.314 & 0.148 & 0.707 & 0.886 & 0.698 & 0.863 & 0.430 & 0.728 & 0.771 & 0.740 & 0.340 & 0.429 & 0.663 & 0.216 & 0.667\\
step by step &  & 0.619 & 0.309 & 0.131 & 0.707 & 0.868 & 0.696 & 0.862 & 0.445 & 0.748 & 0.711 & 0.706 & 0.330 & 0.327 & 0.646 & 0.276 & 0.710\\
openai human written examples &  & 0.630 & 0.302 & 0.165 & 0.707 & 0.888 & 0.723 & 0.854 & 0.410 & 0.762 & 0.695 & 0.740 & 0.343 & 0.416 & 0.679 & 0.252 & 0.707\\
gpt4 style in context examples &  & 0.605 & 0.265 & 0.152 & 0.726 & 0.882 & 0.724 & 0.862 & 0.445 & 0.760 & 0.706 & 0.736 & 0.380 & 0.408 & 0.665 & 0.226 & 0.737\\
rewrite groundtruth in own words &  & 0.497 & 0.241 & 0.139 & 0.700 &  &  &  &  &  &  &  & 0.297 &  &  &  & \\
\hline\hline
gold label & llama 3 instruct &  &  &  & 0.738 & 0.979 & 0.759 & 0.850 & 0.430 & 0.754 & 0.767 & 0.744 &  & 0.510 & 0.769 & 0.342 & 0.643\\
groundtruth &  & 0.677 & 0.408 & 0.241 & 0.706 &  &  &  &  &  &  &  & 0.443 &  &  &  & \\
gpt4 &  & 0.817 & 0.557 & 0.312 & 0.748 & 0.865 & 0.698 & 0.866 & 0.455 & 0.762 & 0.808 & 0.792 & 0.483 & 0.469 & 0.707 & 0.233 & 0.650\\
claude &  & 0.796 & 0.504 & 0.253 & 0.760 & 0.858 & 0.716 & 0.858 & 0.440 & 0.766 & 0.797 & 0.762 & 0.457 & 0.547 & 0.709 & 0.246 & 0.727\\
mini gpt4 &  & 0.810 & 0.548 & 0.274 & 0.719 & 0.863 & 0.664 & 0.871 & 0.430 & 0.751 & 0.811 & 0.810 & 0.487 & 0.384 & 0.727 & 0.226 & 0.633\\
step by step &  & 0.796 & 0.561 & 0.266 & 0.733 & 0.867 & 0.721 & 0.846 & 0.420 & 0.777 & 0.792 & 0.780 & 0.457 & 0.233 & 0.708 & 0.249 & 0.697\\
openai human written examples &  & 0.809 & 0.547 & 0.291 & 0.735 & 0.894 & 0.716 & 0.868 & 0.435 & 0.764 & 0.801 & 0.806 & 0.487 & 0.343 & 0.709 & 0.209 & 0.680\\
gpt4 style in context examples &  & 0.798 & 0.484 & 0.291 & 0.733 & 0.890 & 0.720 & 0.878 & 0.440 & 0.751 & 0.812 & 0.792 & 0.463 & 0.416 & 0.723 & 0.279 & 0.680\\
rewrite groundtruth in own words &  & 0.749 & 0.445 & 0.253 & 0.733 &  &  &  &  &  &  &  & 0.447 &  &  &  & \\
\hline\hline
gold label & qwen &  &  &  & 0.817 & 0.898 & 0.735 & 0.867 & 0.470 & 0.771 & 0.854 & 0.668 &  & 0.514 & 0.737 & 0.306 & 0.613\\
groundtruth &  & 0.896 & 0.892 & 0.658 & 0.798 &  &  &  &  &  &  &  & 0.580 &  &  &  & \\
gpt4 &  & 0.901 & 0.904 & 0.692 & 0.794 & 0.855 & 0.703 & 0.878 & 0.555 & 0.759 & 0.884 & 0.800 & 0.583 & 0.437 & 0.730 & 0.312 & 0.667\\
claude &  & 0.901 & 0.903 & 0.654 & 0.784 & 0.857 & 0.722 & 0.878 & 0.555 & 0.765 & 0.877 & 0.790 & 0.610 & 0.465 & 0.721 & 0.302 & 0.657\\
mini gpt4 &  & 0.903 & 0.904 & 0.662 & 0.789 & 0.872 & 0.716 & 0.882 & 0.565 & 0.765 & 0.891 & 0.822 & 0.647 & 0.371 & 0.700 & 0.312 & 0.667\\
step by step &  & 0.899 & 0.907 & 0.646 & 0.790 & 0.866 & 0.723 & 0.883 & 0.550 & 0.775 & 0.881 & 0.826 & 0.620 & 0.420 & 0.711 & 0.292 & 0.667\\
openai human written examples &  & 0.904 & 0.908 & 0.641 & 0.786 & 0.868 & 0.696 & 0.883 & 0.550 & 0.787 & 0.884 & 0.822 & 0.633 & 0.465 & 0.720 & 0.282 & 0.620\\
gpt4 style in context examples &  & 0.900 & 0.903 & 0.658 & 0.805 & 0.878 & 0.730 & 0.882 & 0.585 & 0.787 & 0.870 & 0.814 & 0.643 & 0.294 & 0.742 & 0.299 & 0.637\\
rewrite groundtruth in own words &  & 0.897 & 0.907 & 0.692 & 0.785 &  &  &  &  &  &  &  & 0.590 &  &  &  & \\
\hline
\end{tabular}}
\caption{seed 1 train datasize 1000 lr 2e-05 epoch num 20}
\label{tab:ntrain_1000_lr_2e-05_seed1}
\end{table*}

\begin{table*}[t!]
\centering
\resizebox{1.0\textwidth}{!}{
\begin{tabular}{l|l|c|c|c|c|c|c|c|c|c|c|c|c|c|c|c|c}
\hline
Data Generation Strategy & Model Type & gsm8k & math algebra & math geometry & ecqa & boolq & winogrande & piqa & agieval & squad & arc challenge & drop & mbpp & api bank & hellaswag & mmlu pro law & mmlu moral scenarios \\ \hline
gold label & mistral &  &  &  & 0.681 & 0.996 & 0.743 & 0.838 & 0.450 & 0.741 & 0.734 & 0.656 &  & 0.449 & 0.776 & 0.269 & 0.663\\
groundtruth &  & 0.441 & 0.211 & 0.101 & 0.679 &  &  &  &  &  &  &  & 0.350 &  &  &  & \\
gpt4 &  & 0.617 & 0.315 & 0.169 & 0.708 & 0.868 & 0.700 & 0.870 & 0.415 & 0.739 & 0.661 & 0.720 & 0.343 & 0.482 & 0.641 & 0.276 & 0.703\\
claude &  & 0.612 & 0.277 & 0.148 & 0.742 & 0.883 & 0.716 & 0.856 & 0.410 & 0.744 & 0.743 & 0.726 & 0.367 & 0.445 & 0.570 & 0.246 & 0.713\\
mini gpt4 &  & 0.622 & 0.320 & 0.177 & 0.703 & 0.865 & 0.688 & 0.855 & 0.435 & 0.740 & 0.768 & 0.708 & 0.353 & 0.429 & 0.670 & 0.219 & 0.697\\
step by step &  & 0.622 & 0.322 & 0.139 & 0.709 & 0.866 & 0.697 & 0.843 & 0.430 & 0.763 & 0.700 & 0.714 & 0.360 & 0.298 & 0.661 & 0.219 & 0.720\\
openai human written examples &  & 0.614 & 0.323 & 0.156 & 0.701 & 0.900 & 0.718 & 0.855 & 0.405 & 0.754 & 0.663 & 0.748 & 0.363 & 0.408 & 0.679 & 0.246 & 0.720\\
gpt4 style in context examples &  & 0.606 & 0.251 & 0.165 & 0.712 & 0.884 & 0.724 & 0.860 & 0.420 & 0.771 & 0.711 & 0.748 & 0.373 & 0.449 & 0.673 & 0.266 & 0.737\\
rewrite groundtruth in own words &  & 0.506 & 0.222 & 0.135 & 0.703 &  &  &  &  &  &  &  & 0.327 &  &  &  & \\
\hline\hline
gold label & llama 3 instruct &  &  &  & 0.735 & 0.980 & 0.760 & 0.865 & 0.445 & 0.757 & 0.762 & 0.740 &  & 0.465 & 0.784 & 0.329 & 0.663\\
groundtruth &  & 0.696 & 0.415 & 0.228 & 0.690 &  &  &  &  &  &  &  & 0.413 &  &  &  & \\
gpt4 &  & 0.806 & 0.553 & 0.278 & 0.733 & 0.864 & 0.697 & 0.865 & 0.450 & 0.742 & 0.824 & 0.748 & 0.487 & 0.445 & 0.725 & 0.233 & 0.683\\
claude &  & 0.789 & 0.489 & 0.257 & 0.734 & 0.866 & 0.685 & 0.846 & 0.450 & 0.759 & 0.800 & 0.770 & 0.507 & 0.547 & 0.716 & 0.243 & 0.743\\
mini gpt4 &  & 0.795 & 0.536 & 0.287 & 0.733 & 0.866 & 0.690 & 0.869 & 0.450 & 0.754 & 0.796 & 0.686 & 0.467 & 0.367 & 0.709 & 0.246 & 0.640\\
step by step &  & 0.800 & 0.551 & 0.245 & 0.719 & 0.884 & 0.707 & 0.865 & 0.460 & 0.767 & 0.782 & 0.792 & 0.467 & 0.245 & 0.697 & 0.269 & 0.653\\
openai human written examples &  & 0.796 & 0.529 & 0.287 & 0.736 & 0.884 & 0.714 & 0.863 & 0.450 & 0.757 & 0.808 & 0.800 & 0.460 & 0.367 & 0.689 & 0.223 & 0.680\\
gpt4 style in context examples &  & 0.800 & 0.500 & 0.283 & 0.729 & 0.876 & 0.709 & 0.856 & 0.440 & 0.767 & 0.809 & 0.816 & 0.480 & 0.433 & 0.708 & 0.252 & 0.683\\
rewrite groundtruth in own words &  & 0.754 & 0.431 & 0.291 & 0.715 &  &  &  &  &  &  &  & 0.457 &  &  &  & \\
\hline\hline
gold label & qwen &  &  &  & 0.818 & 0.887 & 0.724 & 0.867 & 0.495 & 0.774 & 0.861 & 0.652 &  & 0.539 & 0.740 & 0.302 & 0.590\\
groundtruth &  & 0.901 & 0.910 & 0.675 & 0.758 &  &  &  &  &  &  &  & 0.623 &  &  &  & \\
gpt4 &  & 0.897 & 0.892 & 0.654 & 0.791 & 0.858 & 0.710 & 0.883 & 0.540 & 0.777 & 0.882 & 0.788 & 0.603 & 0.433 & 0.703 & 0.306 & 0.607\\
claude &  & 0.881 & 0.916 & 0.641 & 0.785 & 0.859 & 0.735 & 0.877 & 0.540 & 0.762 & 0.872 & 0.798 & 0.597 & 0.461 & 0.732 & 0.292 & 0.690\\
mini gpt4 &  & 0.902 & 0.904 & 0.658 & 0.778 & 0.875 & 0.711 & 0.880 & 0.555 & 0.760 & 0.890 & 0.798 & 0.613 & 0.396 & 0.696 & 0.309 & 0.677\\
step by step &  & 0.886 & 0.907 & 0.679 & 0.770 & 0.859 & 0.715 & 0.869 & 0.560 & 0.767 & 0.867 & 0.792 & 0.597 & 0.404 & 0.711 & 0.332 & 0.677\\
openai human written examples &  & 0.900 & 0.887 & 0.646 & 0.794 & 0.881 & 0.707 & 0.872 & 0.540 & 0.802 & 0.895 & 0.804 & 0.603 & 0.449 & 0.724 & 0.306 & 0.640\\
gpt4 style in context examples &  & 0.911 & 0.908 & 0.650 & 0.792 & 0.875 & 0.728 & 0.891 & 0.535 & 0.790 & 0.866 & 0.824 & 0.577 & 0.318 & 0.734 & 0.316 & 0.643\\
rewrite groundtruth in own words &  & 0.905 & 0.899 & 0.637 & 0.799 &  &  &  &  &  &  &  & 0.600 &  &  &  & \\
\hline
\end{tabular}}
\caption{seed 2 train datasize 1000 lr 2e-05 epoch num 20}
\label{tab:ntrain_1000_lr_2e-05_seed2}
\end{table*}

\begin{table*}[t!]
\centering
\resizebox{1.0\textwidth}{!}{
\begin{tabular}{l|l|c|c|c|c|c|c|c|c|c|c|c|c|c|c|c|c}
\hline
Data Generation Strategy & Model Type & gsm8k & math algebra & math geometry & ecqa & boolq & winogrande & piqa & agieval & squad & arc challenge & drop & mbpp & api bank & hellaswag & mmlu pro law & mmlu moral scenarios \\ \hline
gold label & mistral &  &  &  & 0.627 & 0.869 & 0.608 & 0.814 & 0.430 & 0.582 & 0.704 & 0.482 &  & 0.220 & 0.625 & 0.153 & 0.420\\
groundtruth &  & 0.420 & 0.205 & 0.101 & 0.591 &  &  &  &  &  &  &  & 0.267 &  &  &  & \\
gpt4 &  & 0.513 & 0.231 & 0.101 & 0.596 & 0.837 & 0.636 & 0.790 & 0.345 & 0.333 & 0.624 & 0.244 & 0.317 & 0.249 & 0.269 & 0.166 & 0.380\\
claude &  & 0.505 & 0.215 & 0.110 & 0.634 & 0.837 & 0.627 & 0.804 & 0.400 & 0.290 & 0.630 & 0.250 & 0.340 & 0.257 & 0.284 & 0.179 & 0.413\\
mini gpt4 &  & 0.511 & 0.223 & 0.097 & 0.619 & 0.845 & 0.644 & 0.782 & 0.360 & 0.404 & 0.633 & 0.210 & 0.337 & 0.253 & 0.223 & 0.183 & 0.343\\
step by step &  & 0.494 & 0.247 & 0.080 & 0.593 & 0.845 & 0.636 & 0.765 & 0.355 & 0.314 & 0.618 & 0.092 & 0.317 & 0.265 & 0.254 & 0.183 & 0.403\\
openai human written examples &  & 0.504 & 0.230 & 0.118 & 0.611 & 0.853 & 0.639 & 0.811 & 0.355 & 0.467 & 0.578 & 0.280 & 0.317 & 0.257 & 0.316 & 0.166 & 0.517\\
gpt4 style in context examples &  & 0.500 & 0.245 & 0.114 & 0.560 & 0.845 & 0.649 & 0.789 & 0.340 & 0.312 & 0.611 & 0.124 & 0.337 & 0.208 & 0.295 & 0.183 & 0.423\\
rewrite groundtruth in own words &  & 0.450 & 0.214 & 0.110 & 0.603 &  &  &  &  &  &  &  & 0.317 &  &  &  & \\
\hline\hline
gold label & llama 3 instruct &  &  &  & 0.710 & 0.852 & 0.636 & 0.789 & 0.395 & 0.680 & 0.764 & 0.620 &  & 0.082 & 0.610 & 0.196 & 0.200\\
groundtruth &  & 0.794 & 0.460 & 0.249 & 0.691 &  &  &  &  &  &  &  & 0.407 &  &  &  & \\
gpt4 &  & 0.791 & 0.491 & 0.266 & 0.686 & 0.802 & 0.634 & 0.801 & 0.430 & 0.504 & 0.760 & 0.410 & 0.480 & 0.082 & 0.592 & 0.223 & 0.387\\
claude &  & 0.804 & 0.492 & 0.266 & 0.699 & 0.806 & 0.640 & 0.821 & 0.450 & 0.495 & 0.739 & 0.420 & 0.483 & 0.082 & 0.608 & 0.233 & 0.430\\
mini gpt4 &  & 0.797 & 0.477 & 0.257 & 0.710 & 0.800 & 0.621 & 0.823 & 0.425 & 0.509 & 0.751 & 0.400 & 0.497 & 0.086 & 0.589 & 0.229 & 0.373\\
step by step &  & 0.808 & 0.496 & 0.219 & 0.702 & 0.818 & 0.626 & 0.799 & 0.435 & 0.565 & 0.743 & 0.488 & 0.477 & 0.110 & 0.608 & 0.199 & 0.407\\
openai human written examples &  & 0.809 & 0.472 & 0.257 & 0.720 & 0.810 & 0.630 & 0.815 & 0.440 & 0.564 & 0.747 & 0.414 & 0.500 & 0.078 & 0.600 & 0.236 & 0.387\\
gpt4 style in context examples &  & 0.800 & 0.434 & 0.266 & 0.695 & 0.793 & 0.638 & 0.808 & 0.455 & 0.429 & 0.762 & 0.344 & 0.497 & 0.090 & 0.582 & 0.229 & 0.407\\
rewrite groundtruth in own words &  & 0.813 & 0.480 & 0.262 & 0.718 &  &  &  &  &  &  &  & 0.447 &  &  &  & \\
\hline\hline
gold label & qwen &  &  &  & 0.791 & 0.843 & 0.677 & 0.875 & 0.465 & 0.703 & 0.877 & 0.334 &  & 0.220 & 0.702 & 0.306 & 0.387\\
groundtruth &  & 0.913 & 0.918 & 0.679 & 0.792 &  &  &  &  &  &  &  & 0.603 &  &  &  & \\
gpt4 &  & 0.908 & 0.898 & 0.692 & 0.802 & 0.831 & 0.711 & 0.863 & 0.560 & 0.661 & 0.891 & 0.092 & 0.637 & 0.237 & 0.697 & 0.326 & 0.580\\
claude &  & 0.911 & 0.912 & 0.679 & 0.788 & 0.837 & 0.718 & 0.876 & 0.550 & 0.652 & 0.894 & 0.114 & 0.640 & 0.237 & 0.691 & 0.282 & 0.577\\
mini gpt4 &  & 0.902 & 0.914 & 0.667 & 0.791 & 0.852 & 0.720 & 0.884 & 0.550 & 0.660 & 0.891 & 0.068 & 0.600 & 0.237 & 0.702 & 0.296 & 0.583\\
step by step &  & 0.909 & 0.919 & 0.599 & 0.802 & 0.848 & 0.708 & 0.870 & 0.545 & 0.682 & 0.879 & 0.062 & 0.617 & 0.224 & 0.690 & 0.309 & 0.563\\
openai human written examples &  & 0.900 & 0.918 & 0.671 & 0.789 & 0.842 & 0.718 & 0.864 & 0.545 & 0.681 & 0.887 & 0.090 & 0.597 & 0.237 & 0.706 & 0.322 & 0.563\\
gpt4 style in context examples &  & 0.918 & 0.914 & 0.679 & 0.798 & 0.836 & 0.710 & 0.865 & 0.535 & 0.678 & 0.888 & 0.050 & 0.627 & 0.196 & 0.696 & 0.326 & 0.567\\
rewrite groundtruth in own words &  & 0.910 & 0.916 & 0.667 & 0.782 &  &  &  &  &  &  &  & 0.590 &  &  &  & \\
\hline
\end{tabular}}
\caption{seed 0 train datasize 100 lr 2e-05 epoch num 20}
\label{tab:ntrain_100_lr_2e-05_seed0}
\end{table*}

\begin{table*}[t!]
\centering
\resizebox{1.0\textwidth}{!}{
\begin{tabular}{l|l|c|c|c|c|c|c|c|c|c|c|c|c|c|c|c|c}
\hline
Data Generation Strategy & Model Type & gsm8k & math algebra & math geometry & ecqa & boolq & winogrande & piqa & agieval & squad & arc challenge & drop & mbpp & api bank & hellaswag & mmlu pro law & mmlu moral scenarios \\ \hline
gold label & mistral &  &  &  & 0.681 & 0.870 & 0.694 & 0.830 & 0.420 & 0.730 & 0.726 & 0.620 &  & 0.486 & 0.737 & 0.236 & 0.677\\
groundtruth &  & 0.409 & 0.186 & 0.093 & 0.638 &  &  &  &  &  &  &  & 0.293 &  &  &  & \\
gpt4 &  & 0.586 & 0.270 & 0.152 & 0.672 & 0.864 & 0.686 & 0.821 & 0.455 & 0.649 & 0.736 & 0.670 & 0.340 & 0.404 & 0.623 & 0.233 & 0.687\\
claude &  & 0.554 & 0.237 & 0.122 & 0.663 & 0.858 & 0.701 & 0.855 & 0.405 & 0.690 & 0.760 & 0.662 & 0.360 & 0.400 & 0.619 & 0.243 & 0.710\\
mini gpt4 &  & 0.514 & 0.266 & 0.152 & 0.705 & 0.850 & 0.674 & 0.847 & 0.425 & 0.670 & 0.739 & 0.666 & 0.357 & 0.359 & 0.651 & 0.233 & 0.623\\
step by step &  & 0.575 & 0.235 & 0.131 & 0.662 & 0.853 & 0.667 & 0.842 & 0.415 & 0.691 & 0.746 & 0.646 & 0.327 & 0.286 & 0.575 & 0.233 & 0.593\\
openai human written examples &  & 0.536 & 0.278 & 0.156 & 0.674 & 0.874 & 0.665 & 0.850 & 0.435 & 0.700 & 0.764 & 0.698 & 0.340 & 0.302 & 0.628 & 0.229 & 0.677\\
gpt4 style in context examples &  & 0.548 & 0.222 & 0.156 & 0.658 & 0.879 & 0.681 & 0.864 & 0.430 & 0.676 & 0.741 & 0.650 & 0.333 & 0.343 & 0.628 & 0.203 & 0.687\\
rewrite groundtruth in own words &  & 0.443 & 0.202 & 0.101 &  &  &  &  &  &  &  &  & 0.330 &  &  &  & \\
\hline\hline
gold label & llama 3 instruct &  &  &  & 0.705 & 0.866 & 0.675 & 0.847 & 0.430 & 0.727 & 0.773 & 0.684 &  & 0.494 & 0.682 & 0.299 & 0.633\\
groundtruth &  & 0.683 & 0.404 & 0.211 & 0.679 &  &  &  &  &  &  &  & 0.430 &  &  &  & \\
gpt4 &  & 0.798 & 0.529 & 0.257 & 0.731 & 0.864 & 0.679 & 0.845 & 0.440 & 0.729 & 0.815 & 0.734 & 0.470 & 0.424 & 0.711 & 0.246 & 0.683\\
claude &  & 0.805 & 0.495 & 0.224 & 0.712 & 0.834 & 0.694 & 0.857 & 0.420 & 0.744 & 0.789 & 0.742 & 0.467 &  & 0.677 & 0.226 & 0.693\\
mini gpt4 &  & 0.807 & 0.504 & 0.278 & 0.719 & 0.852 & 0.674 & 0.858 & 0.445 & 0.746 & 0.795 & 0.744 & 0.473 & 0.335 & 0.676 & 0.266 & 0.630\\
step by step &  & 0.779 & 0.528 & 0.198 & 0.690 & 0.874 & 0.683 & 0.863 & 0.435 & 0.736 & 0.797 & 0.708 & 0.457 & 0.253 & 0.688 & 0.233 & 0.620\\
openai human written examples &  & 0.772 & 0.483 & 0.249 & 0.712 & 0.873 & 0.678 & 0.853 & 0.405 & 0.726 & 0.789 & 0.772 & 0.473 & 0.302 & 0.674 & 0.262 & 0.640\\
gpt4 style in context examples &  & 0.794 & 0.488 & 0.283 & 0.712 & 0.861 & 0.690 & 0.859 & 0.440 & 0.729 & 0.770 & 0.754 & 0.473 & 0.380 & 0.702 & 0.259 & 0.693\\
rewrite groundtruth in own words &  & 0.693 & 0.415 & 0.232 &  &  &  &  &  &  &  &  & 0.430 &  &  &  & \\
\hline\hline
gold label & qwen &  &  &  & 0.820 & 0.883 & 0.704 & 0.858 & 0.480 & 0.747 & 0.849 & 0.642 &  & 0.457 & 0.725 & 0.339 & 0.563\\
groundtruth &  & 0.867 & 0.896 & 0.637 & 0.823 &  &  &  &  &  &  &  & 0.523 &  &  &  & \\
gpt4 &  & 0.897 & 0.890 & 0.620 & 0.787 & 0.859 & 0.709 & 0.881 & 0.545 & 0.743 & 0.882 & 0.808 & 0.617 & 0.388 & 0.687 & 0.362 & 0.690\\
claude &  & 0.882 & 0.890 & 0.616 & 0.790 & 0.869 & 0.738 & 0.867 & 0.555 & 0.766 & 0.871 & 0.810 & 0.603 & 0.527 & 0.702 & 0.316 & 0.750\\
mini gpt4 &  & 0.889 & 0.912 & 0.624 & 0.794 & 0.867 & 0.719 & 0.887 & 0.530 & 0.750 & 0.891 & 0.772 & 0.580 & 0.429 & 0.707 & 0.289 & 0.640\\
step by step &  & 0.902 & 0.899 & 0.586 & 0.788 & 0.868 & 0.731 & 0.878 & 0.545 & 0.737 & 0.881 & 0.788 & 0.630 & 0.339 & 0.715 & 0.309 & 0.677\\
openai human written examples &  & 0.892 & 0.899 & 0.616 & 0.783 & 0.874 & 0.727 & 0.883 & 0.565 & 0.776 & 0.875 & 0.824 & 0.590 & 0.392 & 0.694 & 0.233 & 0.643\\
gpt4 style in context examples &  & 0.896 & 0.899 & 0.637 & 0.782 & 0.864 & 0.720 & 0.881 & 0.550 & 0.764 & 0.868 & 0.832 & 0.620 & 0.335 & 0.752 & 0.302 & 0.677\\
rewrite groundtruth in own words &  & 0.899 & 0.892 & 0.646 &  &  &  &  &  &  &  &  & 0.583 &  &  &  & \\
\hline
\end{tabular}}
\caption{seed 0 train datasize 100 lr 0.0002 epoch num 40}
\label{tab:ntrain_100_lr_0.0002_seed0}
\end{table*}

\begin{table*}[ht]
\centering
\resizebox{\textwidth}{!}{
\begin{tabular}{l|l|c|c}
\hline
Benchmark Name & Data Name & Chosen/Not Chosen & Why not chosen \\ \hline
Mistral 7B & Winogrande & $\checkmark$ & \\
 & PIQA & $\checkmark$& \\
 &GSM8K &$\checkmark$ & \\
 & MATH& $\checkmark$& \\
  & MBPP& $\checkmark$& \\
   & MMLU& $\checkmark$& \\
        & AGIEVAL&$\checkmark$ & \\
& ARC Challenge & $\checkmark$& \\
 & BoolQ& $\checkmark$& \\
 & Hellaswag &$\checkmark$ &  \\
  & CommonsenseQA &$\times$ & not a reasoning task\\
    & BBH&  $\times$&  In github, it says this dataset can never used in training.\\
    & SIQA & $\times$ &  not a reasoning task\\
 & OpenbookQA &$\times$ & not a reasoning task \\
 & ARC Easy & $\times$& We already choose ARC Challenge \\
 
 & NaturalQuestions & $\times$& It evaluates world knowledge instead of reasoning ability\\
 & TriviaQA &$\times$ & It evaluates world knowledge instead of reasoning ability \\
 & QuAC&  $\times$& this is a multiturn, muti context qa dataset. evaluation is too hard\\
\hline
Llama 3 & MMLU & $\checkmark$& \\
 & MMLU\_Pro & $\checkmark$ & \\
 & GSM8K & $\checkmark$& \\
 & MATH & $\checkmark$& \\
 & AGIEVAL & $\checkmark$& \\
 & ARC CHALLENGE &$\checkmark$ & \\
 & DROP & $\checkmark$& \\
 & API-BANK & $\checkmark$& \\
& IFEval & $\times$ & less than 650 data \\
 & HumanEval+& $\times$& less than 650 data\\
 & BFCL& $\times$&  (subcategory) less than 650 data\\
 & Nexus& $\times$&  Unable to find the dataset\\
 & GPQA &$\times$ & less than 650 data \\
 & HumanEval &$\times$ & less than 650 data \\
&ZeroSCROLLS/QuALITY&$\times$ & This dataset evaluating model's long context QA ability. The input is too long thus is hard to train. \\
&InfiniteBench/En.MC&$\times$ & This dataset evaluating model's long context QA ability. The input is too long thus is hard to train. \\

&NIH/Multi-needle&$\times$ & Long context QA task. The input is too long thus is hard to train. Llama already achieves 98.8\% accuracy with zero-shot setting.\\

\hline
Qwen2.5 & MMLU & $\checkmark$ & \\
 &MMLU Pro &$\checkmark$ & \\
  & MBPP& $\checkmark$& \\
 & ARC CHALLENGE&$\checkmark$ & \\
  & GSM8K&$\checkmark$ & \\
 &MATH &$\checkmark$ & \\
 & WindoGrande&$\checkmark$ & \\
  & HellaSwag& $\checkmark$&  \\
 &MMLU stem &$\times$ & (subcategory) less than 650 data  \\
 &TruthfulQA &$\times$ & not reasoning task\\
 &GPQA &$\times$ &  less than 650 data\\
  &TheoremQA & $\times$& the data set is tooooo challenging for GPT-4o. it does not have the ability to be a teacher for this task. \\
 &HumanEval & $\times$& less than 650 data \\ 
 & HumanEval+& $\times$& less than 650 data\\
 & MMLU redux&$\times$ & (subcategory) less than 650 data \\
 &BBH & $\times$& In github, it says this dataset can never used in training. \\
 &MBPP+ &$\times$ & less than 650 data \\
 & MultiPL-E& $\times$&  (subcategory) less than 650 data\\

\hline
\end{tabular}
}
\caption{This table explains which data from the Mistral, LLaMA3, and Qwen benchmarks were chosen and why some data were not selected. Multi-lingual dataset is not listed in this Table since our experiment only covers English-only datasets. API-BANK is in Table 16 from Llama 3 technical report.}
\label{tab:why_each_data_is_chosen}
\end{table*}

\begin{table*}[t!]
\centering
\resizebox{1.0\textwidth}{!}{
\begin{tabular}{l|l|c|c|c|c|c|c}
\hline
Data Generation Strategy & Model Type & plan bench generation & plan bench optimality & plan bench generalization & plan bench replaning & plan bench reuse & plan bench verification \\ \hline
gold label & mistral & 0.607 & 0.448 & 0.807 & 0.815 & 0.945 & 0.318\\
groundtruth &  &  &  &  &  &  & \\
gpt4 &  & 0.458 & 0.37 & 0.215 & 0.2 & 0.61 & 0.518\\
claude &  & 0.505 & 0.45 & 0.362 & 0.445 & 0.755 & 0.586\\
mini gpt4 &  & 0.445 & 0.38 & 0.282 & 0.235 & 0.51 & 0.5\\
step by step &  & 0.448 & 0.253 & 0.253 & 0.177 & 0.63 & 0.492\\
openai human written examples &  & 0.485 & 0.43 & 0.302 & 0.28 & 0.73 & 0.474\\
gpt4 style in context examples &  & 0.37 & 0.232 & 0.43 & 0.207 & 0.665 & 0.56\\
\hline\hline
gold label & llama 3 instruct & 0.478 & 0.38 & 0.853 & 0.87 & 0.87 & 0.482\\
groundtruth &  &  &  &  &  &  & \\
gpt4 &  & 0.463 & 0.36 & 0.265 & 0.225 & 0.72 & 0.49\\
claude &  & 0.545 & 0.485 & 0.407 & 0.425 & 0.637 & 0.532\\
mini gpt4 &  & 0.415 & 0.375 & 0.25 & 0.215 & 0.5 & 0.674\\
step by step &  & 0.475 & 0.39 & 0.287 & 0.232 & 0.732 & 0.564\\
openai human written examples &  & 0.458 & 0.372 & 0.307 & 0.25 & 0.603 & 0.516\\
gpt4 style in context examples &  & 0.45 & 0.435 & 0.438 & 0.223 & 0.75 & 0.468\\
\hline\hline
gold label & qwen & 0.32 & 0.285 & 0.38 & 0.732 & 0.845 & 0.516\\
groundtruth &  &  &  &  &  &  & \\
gpt4 &  & 0.237 & 0.253 & 0.145 & 0.2 & 0.515 & 0.538\\
claude &  & 0.325 & 0.265 & 0.277 & 0.46 & 0.393 & 0.534\\
mini gpt4 &  & 0.253 & 0.255 & 0.168 & 0.188 & 0.42 & 0.55\\
step by step &  & 0.265 & 0.182 & 0.135 & 0.17 & 0.577 & 0.536\\
openai human written examples &  & 0.24 & 0.19 & 0.222 & 0.155 & 0.47 & 0.52\\
gpt4 style in context examples &  & 0.302 & 0.268 & 0.338 & 0.2 & 0.597 & 0.534\\
\hline
\end{tabular}}
\caption{seed average of seed\_0,1 train datasize 1000 lr 2e-05 epoch num 40}
\label{tab:ntrain_1000_lr_2e-05_seedaverage_of_seed_0,1_pan_bench}
\end{table*}

\subsection{License of the Dtasets}
All dataset we use are publicly available dataset for research purpose. 
API-BANK (Lv 3 problems)~\citep{li2023api}: CC-BY-SA
GSM8K \citep{cobbe2021training}: MIT license
PIQA \citep{bisk2020piqa}: unkown
BoolQ \citep{clark2019boolq}:CC BY-SA 3.0
MBPP \citep{austin2021program}:CC BY 4.0
DROP \citep{dua2019dropreadingcomprehensionbenchmark}: CC BY-SA 4.0
ARC-Challenge \citep{clark2018think}:CC BY-SA 4.0
PlanBench~\citep{valmeekam2023planbench}: MIT license
MATH (Algebra) and MATH (Geometry) \citep{hendrycks2021measuring}: MIT license
SQuAD \citep{rajpurkar-etal-2016-squad}:SA 4.0 license
MMLU\citep{hendrycks2020measuring}: MIT license
WinoGrande \citep{sakaguchi2021winogrande}: Apache-2.0 license
Hellaswag \citep{zellers-etal-2019-hellaswag}: MIT license  
ECQA \citep{aggarwal2021explanations}: Apache-2.0 license
MMLU\_PRO \citep{wang2024mmlu}:  Apache-2.0 license
AGIEval\citep{zhong2023agieval}: MIT license

\end{document}